%% file: tip.tex
\documentclass[journal]{IEEEtran}

\usepackage{epsfig}
\usepackage{graphicx}
\usepackage{amsmath}
\usepackage{amssymb}

\usepackage{xcolor}
\usepackage{soul}
\usepackage[utf8]{inputenc}
\usepackage[pagebackref=true,breaklinks=true,letterpaper=true,colorlinks,bookmarks=false,citecolor=cyan]{hyperref}

\usepackage[figuresright]{rotating}
\usepackage{textcomp,subfigure,algorithm,algorithmic,multirow,upgreek,tabularx}
\usepackage{graphics,gensymb}
\usepackage{bm,cases,threeparttable}

\usepackage{setspace}

\graphicspath{{Figures/}}
\usepackage{booktabs}

\usepackage{amsfonts}
 
\usepackage{caption}
\ifCLASSINFOpdf
\else
\fi

\begin{document}
%
\title{Layout-to-Image Translation with Double Pooling Generative Adversarial Networks}


\author{\IEEEauthorblockN{Hao Tang and
		Nicu Sebe
}
\thanks{Hao Tang is with the Department of Information Technology and Electrical Engineering, ETH Zurich,  Zurich 8092, Switzerland. E-mail: hao.tang@vision.ee.ethz.ch
\par Nicu Sebe is with the Department of Information Engineering and Computer Science (DISI), University of Trento, Trento 38123, Italy. E-mail: sebe@disi.unitn.it.
\par Corresponding author: Hao Tang.
}}

\markboth{IEEE Transactions on Image Processing}%
{Shell \MakeLowercase{\textit{et al.}}: Bare Demo of IEEEtran.cls for IEEE Transactions on Magnetics Journals}
%

\IEEEtitleabstractindextext{%

\input{0Abstract}

\begin{IEEEkeywords}
GANs, Pooling, Layout-to-Image Translation
\end{IEEEkeywords}}

\maketitle

\IEEEdisplaynontitleabstractindextext

%
\IEEEpeerreviewmaketitle

\input{1Introduction}

\input{2RelatedWork}
\input{3Method}

\input{4Experiments}

\input{5Conclusion}

\section*{Acknowledgments}
This work was supported by the EU H2020 AI4Media No. 951911 project, by the Italy-China collaboration project TALENT:2018YFE0118400, and by the PRIN project PREVUE.

\small
\bibliographystyle{IEEEtran}
\bibliography{reference}

\begin{IEEEbiography}[{\includegraphics[width=1in,height=1.25in,clip,keepaspectratio]{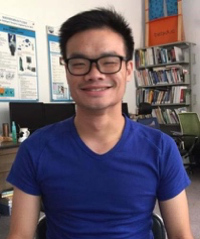}}]{Hao Tang}
is currently a Postdoctoral with Computer Vision Lab, ETH Zurich, Switzerland. 
He received the master’s degree from the School of Electronics and Computer Engineering, Peking University, China and the Ph.D. degree from Multimedia and Human Understanding Group, University of Trento, Italy.
He was a visiting scholar in the Department of Engineering Science at the University of Oxford. His research interests are deep learning, machine learning, and their applications to computer vision.

\end{IEEEbiography}

\begin{IEEEbiography}[{\includegraphics[width=1in,height=1.25in,clip,keepaspectratio]{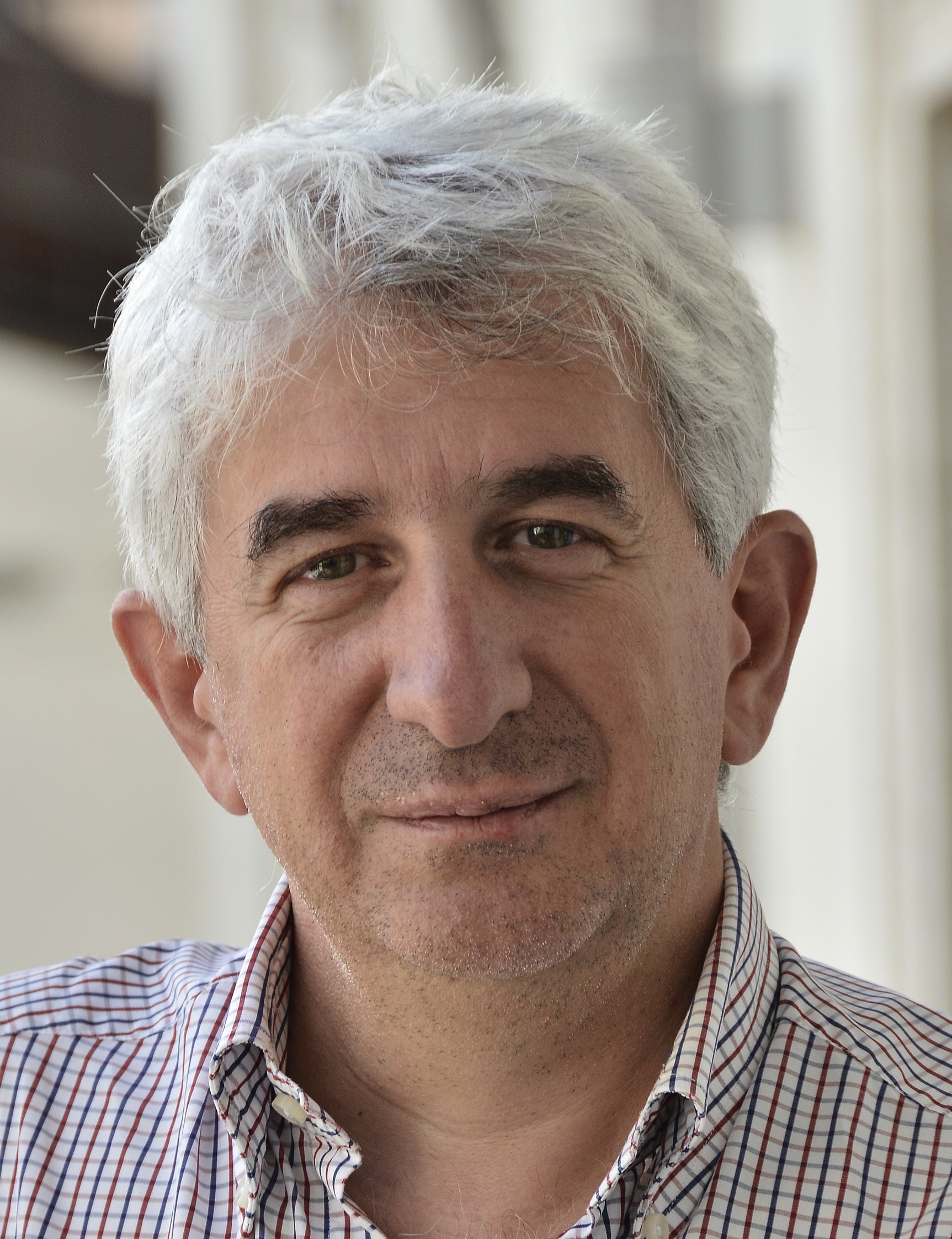}}]{Nicu Sebe} 
is Professor in the University of Trento, Italy, where he is leading the research in the areas of multimedia analysis and human behavior understanding. He was the General Co-Chair of the IEEE FG 2008 and ACM Multimedia 2013.  He was a program chair of ACM Multimedia 2011 and 2007, ECCV 2016, ICCV 2017 and ICPR 2020.  He is a general chair of ACM Multimedia 2022 and a program chair of ECCV 2024. He is a fellow of IAPR.
\end{IEEEbiography}

%
%
%
%
%




\end{document}

%% file: 0Abstract.tex
\begin{abstract}
In this paper, we address the task of layout-to-image translation, which aims to translate an input semantic layout to a realistic image.
One open challenge widely observed in existing methods is the lack of effective semantic constraints during the image translation process, leading to models that cannot preserve the semantic information and ignore the semantic dependencies within the same object.
To address this issue, we propose a novel Double Pooing GAN (DPGAN) for generating photo-realistic and semantically-consistent results from the input layout.
We also propose a novel Double Pooling Module (DPM), which consists of the Square-shape Pooling Module (SPM) and the Rectangle-shape Pooling Module (RPM). 
Specifically, SPM aims to capture short-range semantic dependencies of the input layout with different spatial scales, while RPM aims to capture long-range semantic dependencies from both horizontal and vertical directions.
We then effectively fuse both outputs of SPM and RPM to further enlarge the receptive field of our generator.
Extensive experiments on five popular datasets show that the proposed DPGAN achieves better results than state-of-the-art methods. 
Finally, both SPM and SPM are general and can be seamlessly integrated into any GAN-based architectures to strengthen the feature representation. The code is available at~\url{https://github.com/Ha0Tang/DPGAN}.
\end{abstract}

%% file: 1Introduction.tex
\section{Introduction}
\begin{figure}[t!]
\centering
\subfigure[Answer: \_\_\_\_\_\_\_\_]{\label{fig:a}\includegraphics[width=0.32\linewidth]{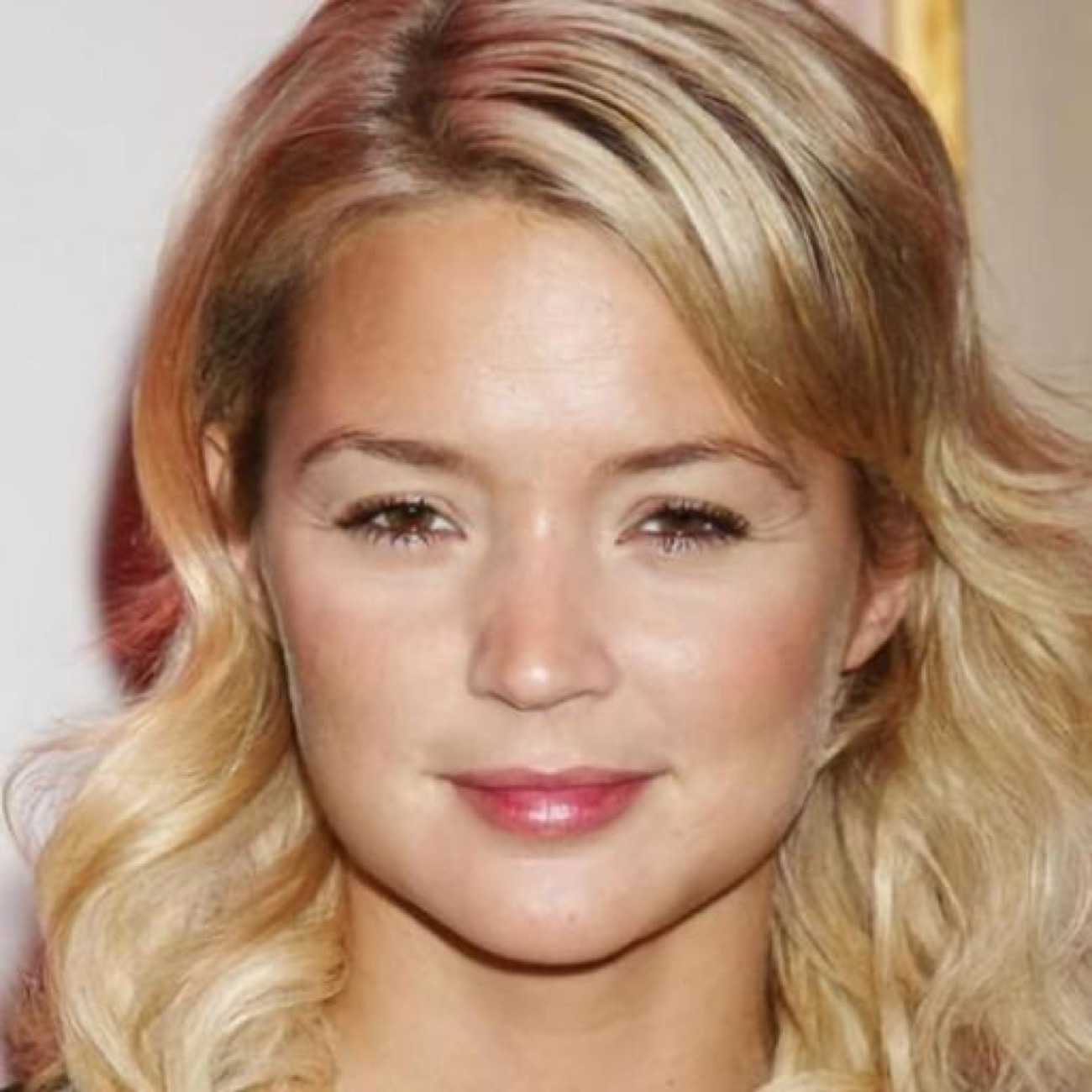}}
\subfigure[Answer: \_\_\_\_\_\_\_\_]{\label{fig:b}\includegraphics[width=0.32\linewidth]{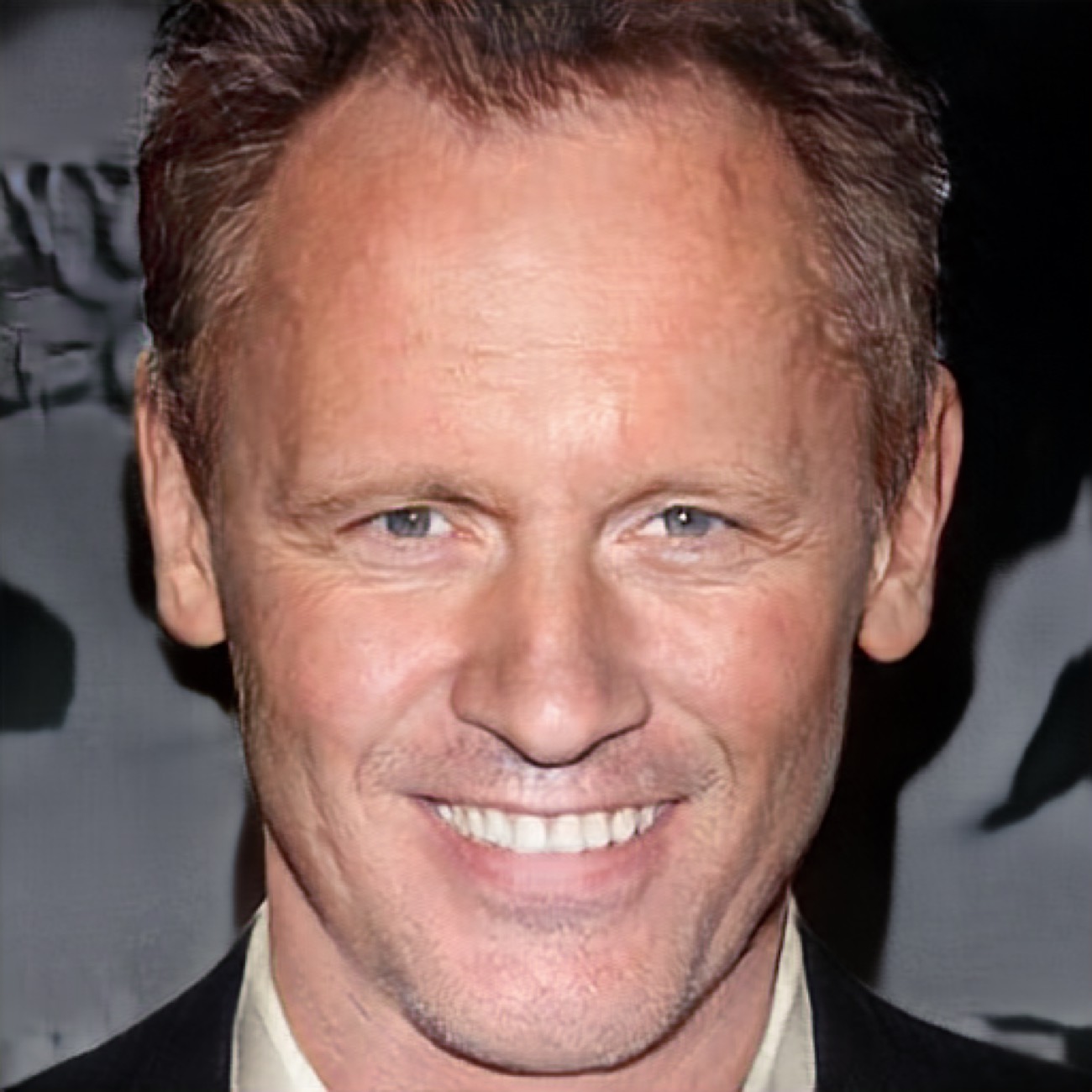}}
\subfigure[Answer: \_\_\_\_\_\_\_\_]{\label{fig:c}\includegraphics[width=0.32\linewidth]{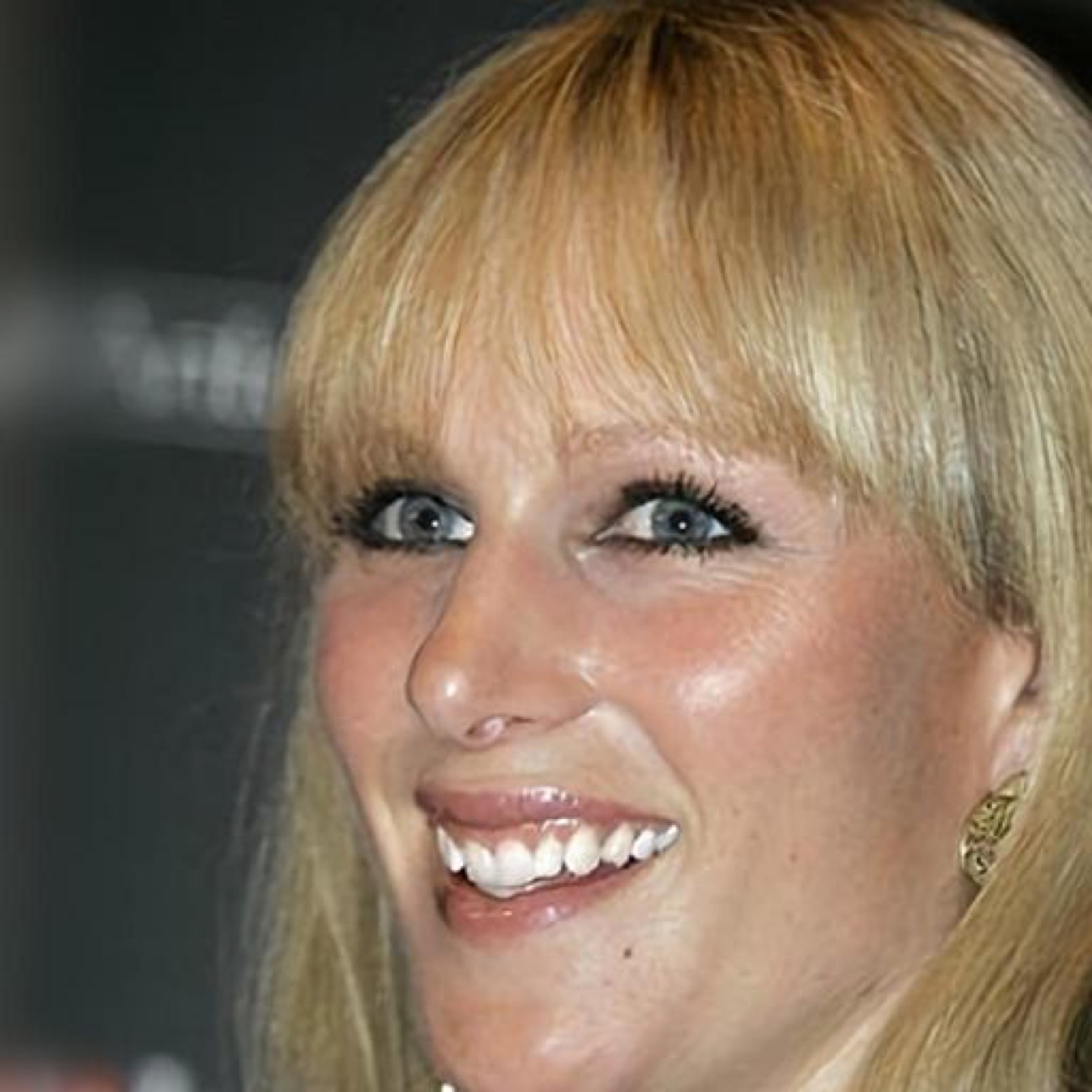}}
\subfigure[Answer: \_\_\_\_\_\_\_\_]{\label{fig:d}\includegraphics[width=0.32\linewidth]{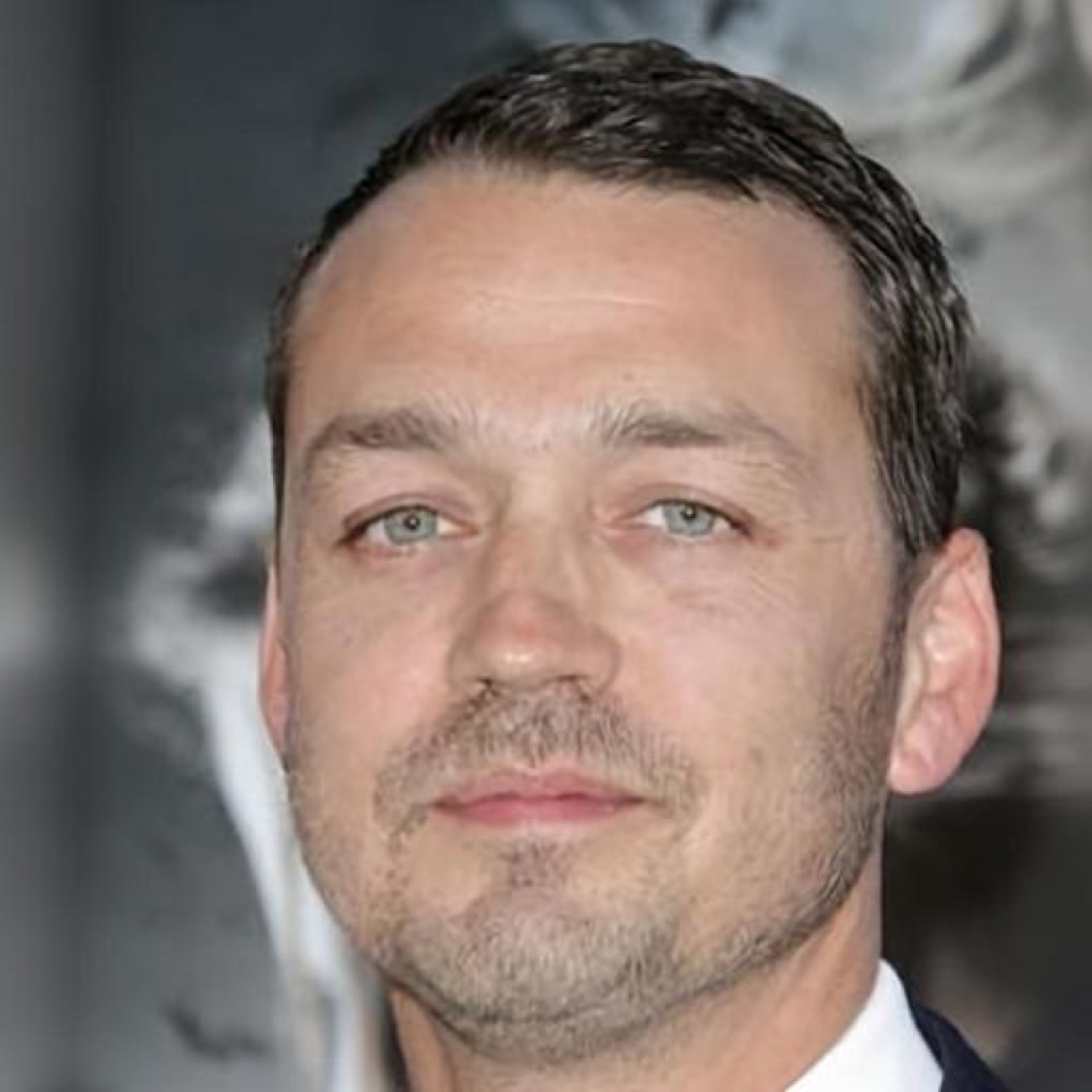}}
\subfigure[Answer: \_\_\_\_\_\_\_\_]{\label{fig:e}\includegraphics[width=0.32\linewidth]{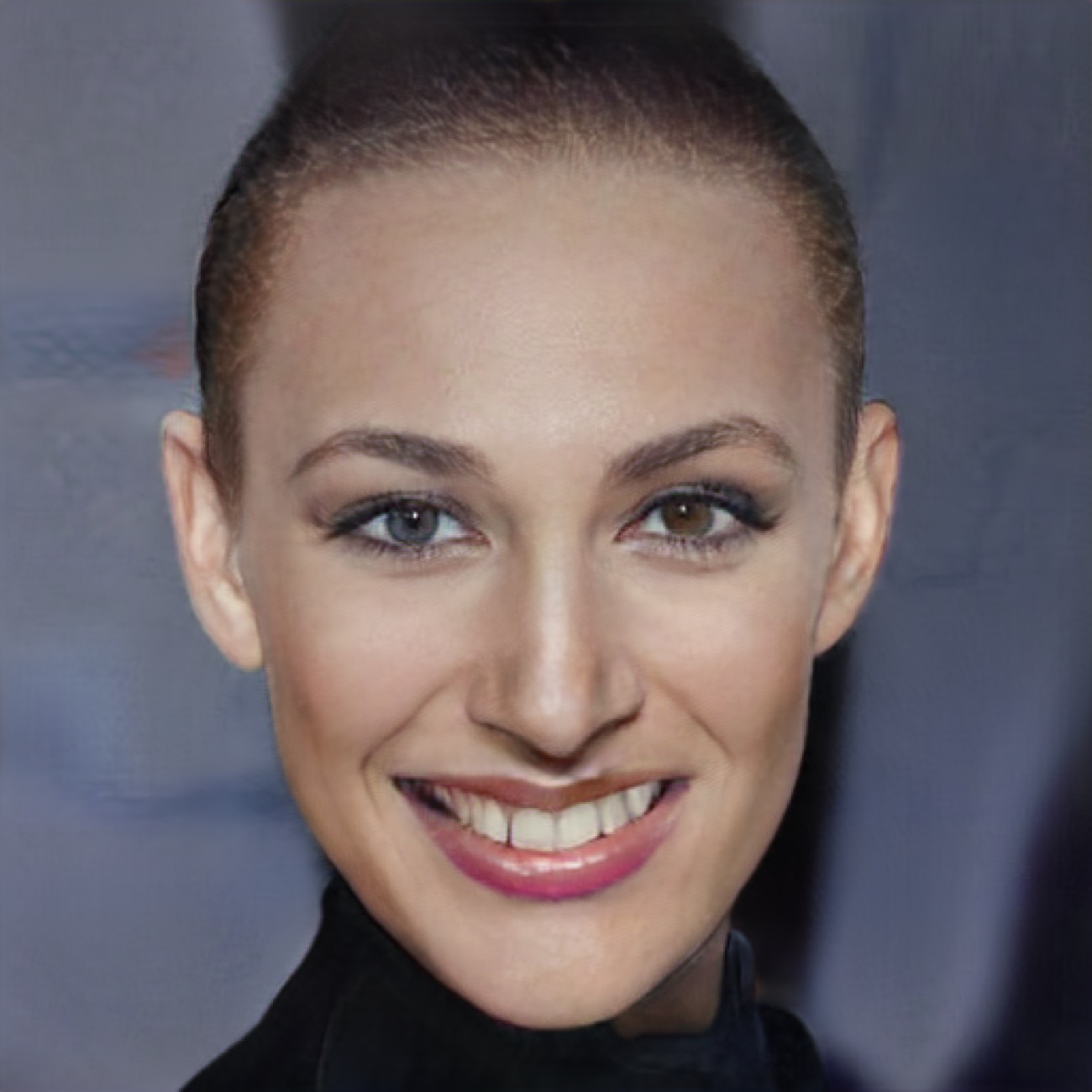}}
\subfigure[Answer: \_\_\_\_\_\_\_\_]{\label{fig:f}\includegraphics[width=0.32\linewidth]{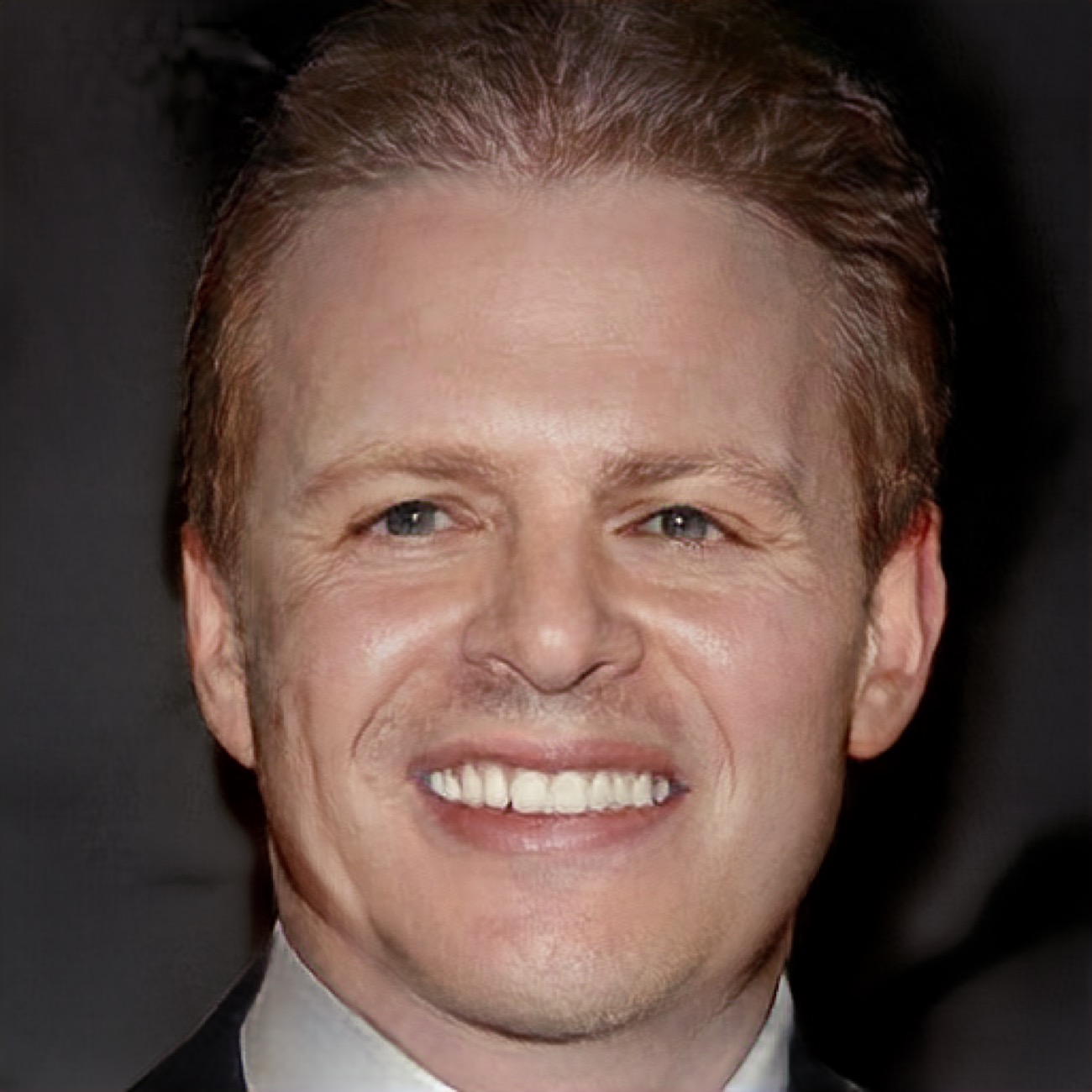}}
\caption{`Real vs. Fake' game: Can you guess which image is real and which has been generated by the proposed DPGAN?}
\label{fig:first}
\end{figure}

In Figure~\ref{fig:first} we show a `Real vs. Fake' game, in which a mix of `real' images are collected from the real world and `fake' images are generated by our GAN model. The goal is to guess which image is real and which one has been generated by the proposed GAN model.
Now you can check your answers below\footnote{Answers: (a) real; (b) fake; (c) real; (d) real; (e) fake; (f) fake.}.
This should be a very challenging and difficult task, considering the recent progress in Generative Adversarial Networks (GANs) \cite{goodfellow2014generative}.

\begin{figure*}[!t]
	\centering
	\includegraphics[width=0.96\linewidth]{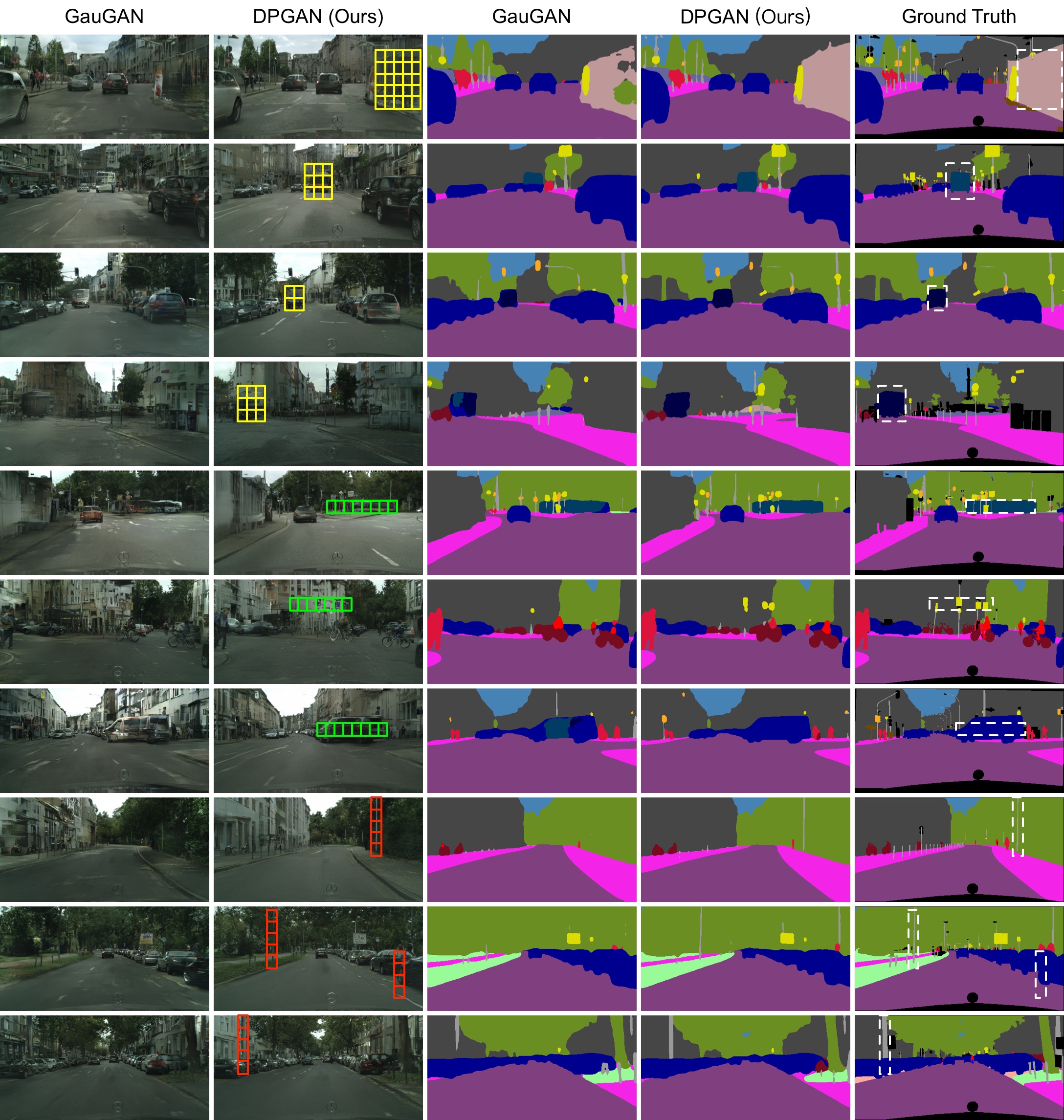}
	\caption{Visualization of our generated semantic maps compared with those from GauGAN~\cite{park2019semantic} on Cityscapes. 
		The proposed SPM (yellow grids) captures short-range and local dependencies, while the proposed RPM (i.e., HRPM and VRPM) captures long-range and global semantic correlations from both horizontal direction (green grids) and vertical direction (red grids), respectively.
		Equipped with both modules, the proposed DPGAN can enlarge the receptive field, thus improves the intra-object semantic consistency.
        Most improved regions are highlighted in the ground truths with white dash boxes.}
	\label{fig:seg}
\end{figure*}

In this paper, we aim to address the challenging layout-to-image translation task, which has a wide range
of real-world applications such as content generation and image editing \cite{chen2017photographic,isola2017image,wang2018high}.
This task has been widely investigated in recent years \cite{wang2018high,park2019semantic,liu2019learning,dundar2020panoptic,jiang2020tsit,tang2020local}.
For example, 
Park et al. \cite{park2019semantic} proposed the GauGAN model with a novel spatially-adaptive normalization to generate realistic images from semantic layouts. 
Tang et al. \cite{tang2020local} proposed the LGGAN framework with a novel local generator for generating realistic small objects and detailed local texture.
Despite the interesting exploration of these methods, we can still observe blurriness and artifacts in their generated results because the existing methods lack an effective semantic dependency modeling to maintain the semantic information of the input layout, causing intra-object semantic inconsistencies such as the fence, buses, and pole generated by GauGAN in Figure~\ref{fig:seg}.

To solve this limitation, we propose a novel Double Pooling GAN (DPGAN) and a novel Double Pooling Module (DPM).
The proposed DPM consists of two sub-modules, i.e., Square-shape Pooling Module (SPM) and Rectangle-shape Pooling Module (RPM).
In particular, SPM aims to capture short-range and local semantic dependencies, leading pixels within the same object to be correlated. Simultaneously, RPM aims to capture long-range and global semantic dependencies from both horizontal and vertical directions.
Finally, we propose seven image-level and feature-level fusion strategies to effectively combine the outputs of both SPM and RPM for generating high-quality and semantically-consistent images.

\begin{figure*}[t]
	\centering
	\includegraphics[width=1\textwidth]{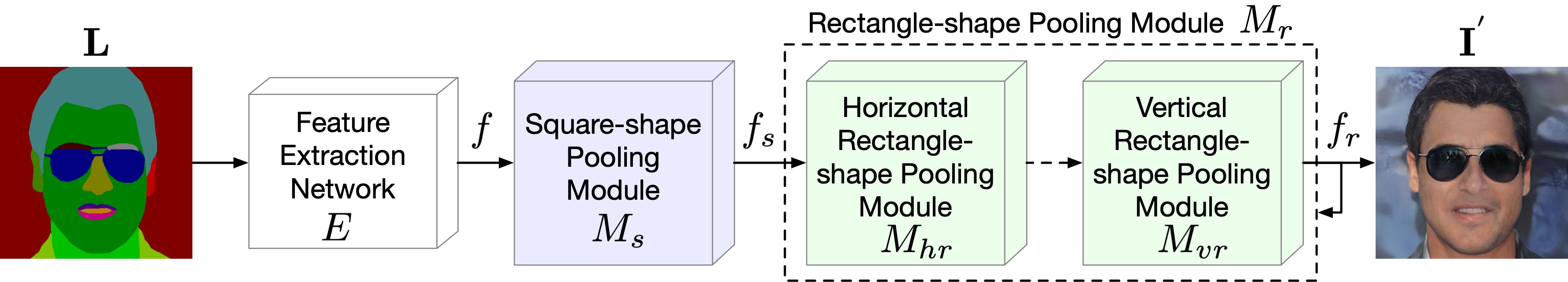}
	\caption{Overview of the generator $G$ of our proposed DPGAN, which consists of a feature extraction network $E$, a Square-shape Pooling Module $M_{s}$, and a Rectangle-shape Pooling Module $M_{r}$.
	All components are trained in an end-to-end fashion so that $M_{s}$ and $M_{r}$ can benefit from each other by capturing both long-range and short-range semantic dependencies.
	}
	\label{fig:method}
\end{figure*}

Overall, the contributions of our paper are:
\begin{itemize}
\item We propose a novel Double Pooing GAN (DPGAN) for the challenging task of layout-to-image translation, which can effectively capture semantic dependencies among different locations of the input layout for generating photo-realistic and semantically-consistent images.
\item We design a novel Double Pooling Module (DPM), which consists of the Square-shape Pooling Module (SPM) and the Rectangle-shape Pooling Module (RPM).
SPM aims at capturing short-range and local semantic dependencies, while RPM aims at modeling long-range and global semantic dependencies from both horizontal and vertical directions.
Both SPM and RPM are general and can be readily applied to existing GAN-based frameworks without modifying the architecture of the network.
\item We conduct extensive experiments on five popular datasets with different image resolutions, i.e., ADE20K \cite{zhou2017scene}, DeepFashion \cite{liu2016deepfashion}, Cityscapes \cite{cordts2016cityscapes}, CelebAMask-HQ \cite{CelebAMask-HQ}, and Facades \cite{tylevcek2013spatial}. 
Both qualitative and quantitative results demonstrate that the proposed DPGAN is able to produce better results than state-of-the-art approaches.
\end{itemize}

%% file: 2RelatedWork.tex
\section{Related Work}

\noindent \textbf{Generative Adversarial Networks (GANs)}
\cite{goodfellow2014generative} are widely used techniques to generate high-quality images \cite{zhang2020dual,karras2019style,tang2020unified,liu2020exocentric,tang2021total,tang2021total}, videos \cite{liu2021cross}, and 3D objects \cite{chen2021unsupervised}.
A GAN framework contains a generator and a discriminator, where the generator tries to generate realistic images to fool the discriminator while the discriminator aims to accurately tell whether an image is real or fake.
Furthermore, Mirza and Osindero propose Conditional GANs (CGANs) \cite{mirza2014conditional} based on GANs by incorporating extra guidance information to generate user-specific images, e.g., category labels \cite{choi2018stargan,tang2019expression,tang2019attribute}, text descriptions \cite{zhang2017stackgan,tao2020df}, human pose/gesture \cite{tang2020xinggan,tang2019cycle,chan2019everybody,tang2020bipartite,tang2018gesturegan}, attention maps \cite{tang2019attention,tang2019multi,duan2021cascade,tang2021attentiongan}.

\noindent \textbf{Layout-to-Image Translation}
aims to turn semantic layouts into realistic images \cite{park2019semantic,liu2019learning,jiang2020tsit,tang2020local,zhu2020sean,ntavelis2020sesame,zhu2020semantically,tang2020dual,tang2020edge}.
For example, Park et al.~\cite{park2019semantic} proposed GauGAN with a novel spatially-adaptive normalization to generate realistic images. 
Although GauGAN \cite{park2019semantic} has achieved promising results, we still observe unsatisfactory aspects mainly in the generated scene details and intra-object completions (see Figure~\ref{fig:seg}), which we believe are mainly due to the lack of short-range and long-range semantic constrains in the input layout. 
The proposed SPM and RPM explicitly address this problem.

Pooling operations are commonly used in semantic segmentation tasks \cite{zhao2017pyramid,he2019adaptive,hou2020strip,huang2019ccnet} to improve the receptive field.
For example, Zhao et al. \cite{zhao2017pyramid} proposed a pyramid pooling module to capture the global context information in the scene parsing task.
Hou et al. \cite{hou2020strip} proposed a new
strip pooling, which considers a long but narrow kernel, for the scene parsing task.
However, to the best of our knowledge, our idea of using pooling modules to capture both short-range and long-range semantic dependencies has not been investigated by any existing layout-to-image translation or even GAN-based image generation approaches.

%% file: 3Method.tex
\section{Double Pooling GANs}

\noindent \textbf{Overview.}
We start by presenting the details of the proposed Dual Pooling GANs (DPGAN), which consists of a generator $G$ and discriminator $D$.
An illustration of the proposed generator $G$ is shown in Figure~\ref{fig:method}, which mainly consists of three components, i.e., a feature extraction network $E$ extracting deep features from the input layout $\mathbf{L}$, a Square-shape Pooling Module (SPM) modeling short-range and local semantic dependencies, and a Rectangle-shape Pooling Module (RPM) capturing long-range and global semantic dependencies from both horizontal and vertical directions. 
SPM and RPM together form our proposed Double Pooling Module (DPM).
Moreover, we propose seven image-level and feature-level fusion methods to combine both the outputs of SPM and RPM.
 
\noindent \textbf{Feature Extraction Network.}
As shown in Figure~\ref{fig:method}, the network $E$ receives the semantic layout $\mathbf{L}$ as input and outputs the deep feature $f$, which can be formulated as, 
\begin{equation}
\begin{aligned}
f = E(\mathbf{L}).
\end{aligned}
\end{equation}
Then,  $f$ is fed into the proposed SPM and RPM for learning short-range and long-rang semantic dependencies, respectively. 

\begin{figure}[t] \small
	\centering
	\includegraphics[width=1\linewidth]{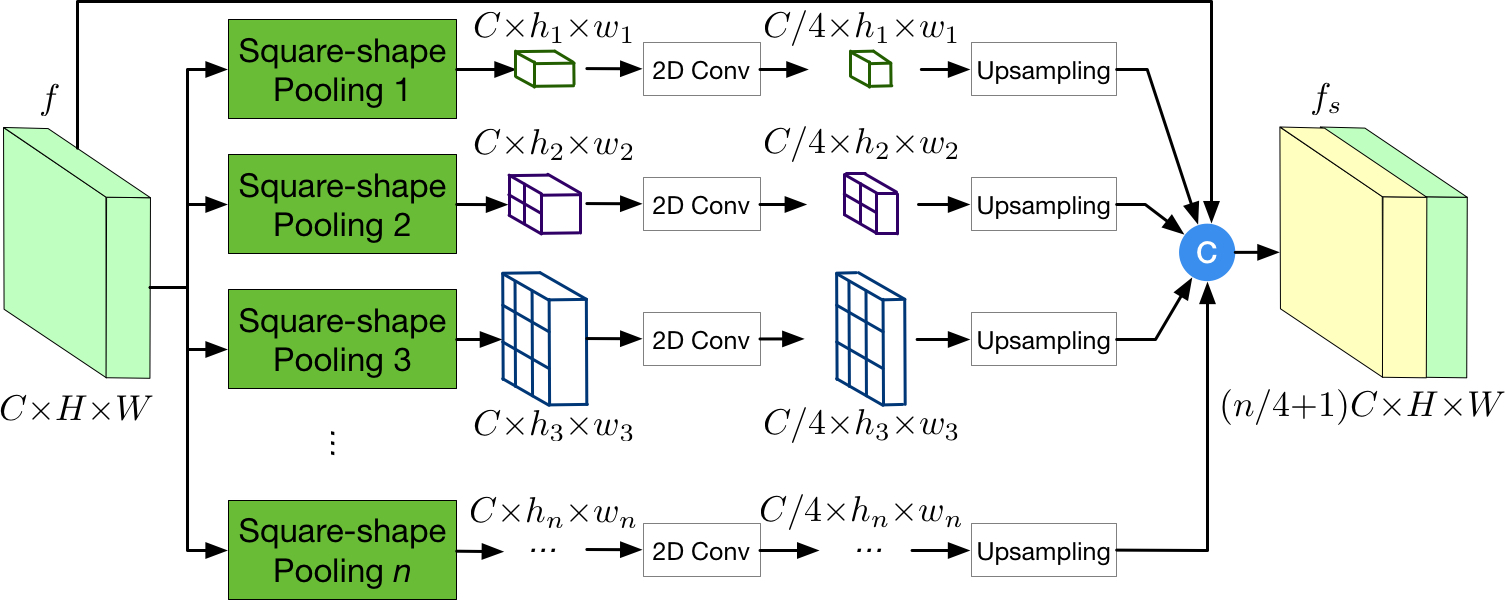}
	\caption{The proposed Square-shape Pooling Module (SPM) which aims to capture short-range and local semantic dependencies. Our SPM is a $n$-level pooling module with different square-kernel size, i.e., ($h_1$, $w_1$), ($h_2$, $w_2$), $\cdots$, ($h_n$, $w_n$), where $\{h_i {=} w_i\}_{i=1}^n$. The symbol $\textcircled{c}$ denotes channel-wise concatenation.}
	\label{fig:spm}
\end{figure}

\begin{figure*}[t]
	\centering
	\includegraphics[width=1\linewidth]{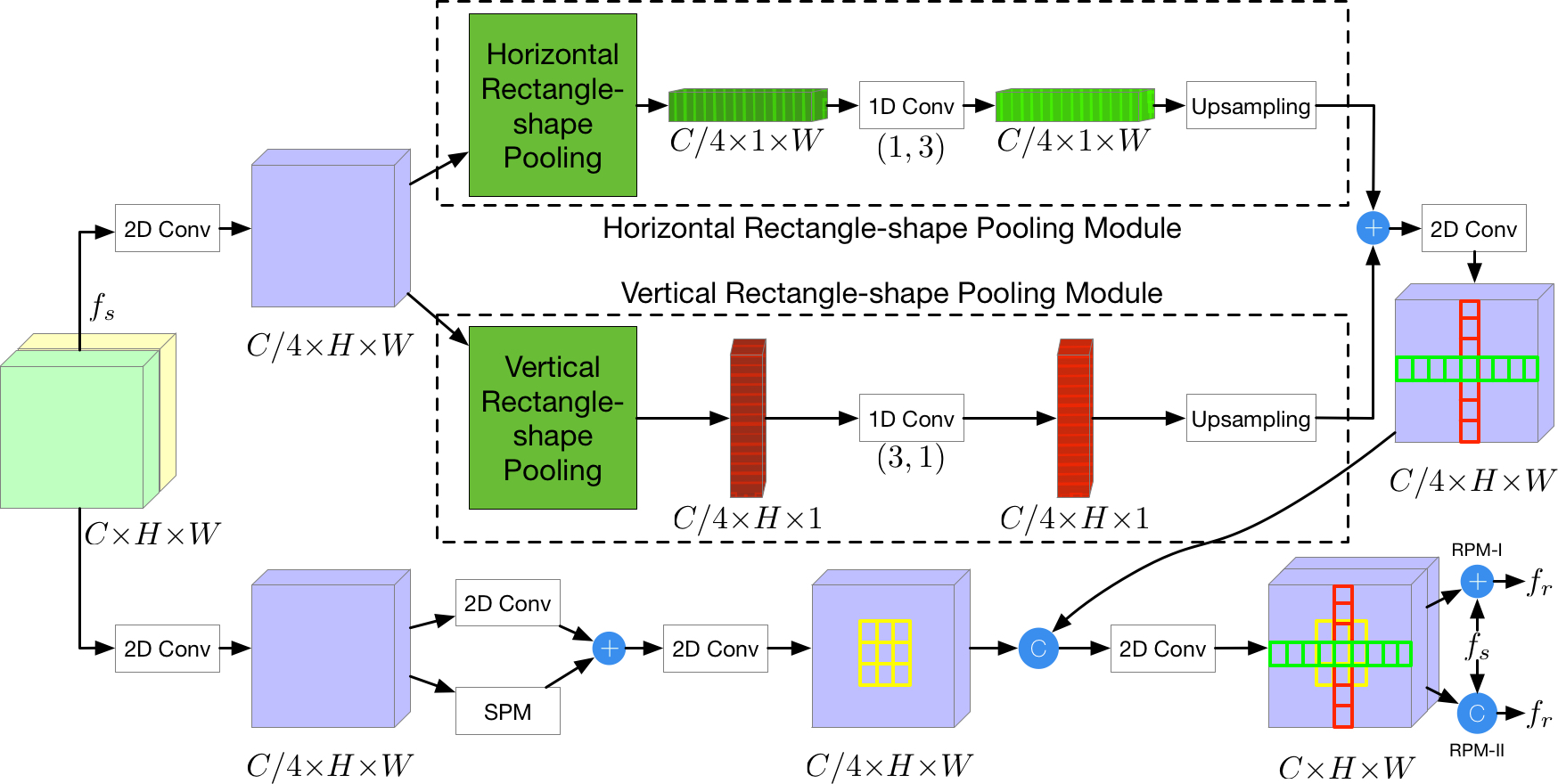}
	\caption{The proposed Rectangle-shape Pooling Module (RPM) which consists of a Horizontal Rectangle-shape Pooling Module (HRPM) and a Vertical Rectangle-shape Pooling Module (VRPM), aiming to capture long-range and global semantic dependencies from horizontal and vertical direction, respectively. The yellow, green, and red grids represent short-dependency, horizontal long-dependency, and vertical long-dependency, respectively. The symbols $\oplus$, and $\textcircled{c}$ denote element-wise addition, and channel-wise concatenation, respectively.}
	\label{fig:rpm}
\end{figure*}

\noindent \textbf{Square-Shape Pooling Module.}
Existing layout-to-image translation methods such as \cite{park2019semantic,liu2019learning,tang2020local,tang2020dual} directly use deep features generated by convolutional operations, leading to limited effective fields-of-views and thus generating different textures in the pixels with the same label.
To model short-range and local semantic dependencies over the deep feature $f$, we propose a Square-shape Pooling Module (SPM).
Note that the idea of the proposed SPM is inspired by the pyramid pooling module proposed in \cite{zhao2017pyramid} and we extend the original module used in image segmentation to a completely different image generation task.

The framework of SPM is elaborated in Figure~\ref{fig:spm}.
Specifically, we first separately feed $f$ into $n$ square-shape pooling layers to produce $n$ new feature maps with different spatial scales.
Consider the $n$-th pooing layer in Figure~\ref{fig:spm}, whose input is the deep feature $f {\in} \mathbb{R}^{C{\times} H {\times} W}$.
The output is the pooled feature map $f^n {\in} \mathbb{R}^{C{\times} h_n {\times} w_n}$, where ($h_n$, $w_n$) is the targeted output size of the $n$-th pooling layer.
We then feed the pooled feature $f^n$ through a convolutional layer for reducing the number of channels, leading to a new feature map $\hat{f^n} {\in} \mathbb{R}^{C/4 {\times} h_n {\times} w_n}$.
After that, we perform an upsampling operation on $\hat{f^n}$ to obtain the feature map $\tilde{f^n}$, which has the same spatial size with $f$.
Mathematically,
\begin{equation}
\begin{aligned}
\tilde{f^n} = {\rm up}_n({\rm conv}({\rm pl}_n(f))),
\end{aligned}
\end{equation}
where ${\rm up}(\cdot)$, ${\rm conv}(\cdot)$ and ${\rm pl}(\cdot)$ denote upsampling, convolutional layer, and square-pooing layer, respectively.

Next, we concatenate all $n$ learned feature maps and the input feature $f$ to produce the final feature $f_s {\in} \mathbb{R}^{(n/4+1)C {\times} H {\times} W}$.
The computation process can be expressed as follow,
\begin{equation}
\begin{aligned}
f_s = {\rm concat}(\tilde{f^1}, \tilde{f^2}, \tilde{f^3}, \cdots, \tilde{f^n}, f),
\end{aligned}
\end{equation}
where ${\rm concat}(\cdot)$ denotes channel-wise concatenation.
By doing so, the feature $f_s$ has short-range and local semantic dependencies with different spatial scales.
Therefore, the pixels with the same semantic label can achieve mutual gains, thus we are improving intra-object semantic consistency (see the first to fourth rows in Figure~\ref{fig:seg}).

\noindent \textbf{Rectangle-Shape Pooling Module.}
The proposed SPM captures only short-range semantic dependencies, as shown in Figure~\ref{fig:seg}.
To capture long-range and global semantic dependencies, we can increase the kernel size of the square pooling.
However, this inevitably incorporates lots of irrelevant regions when processing rectangle-shaped and narrow objects such as the bus and pole shown in Figure~\ref{fig:seg}.

To alleviate this limitation, we propose a novel Rectangle-shape Pooling Module (RPM), which aims to capture long-range and global semantic dependencies from both horizontal and vertical directions.
The idea of the proposed RPM is inspired by the strip pooling module proposed in \cite{hou2020strip} and the framework of RPM is illustrated in Figure~\ref{fig:rpm}. It consists of a Horizontal Rectangle-shape Pooling Module (HRPM) and a Vertical Rectangle-shape Pooling Module (VRPM).
HRPM captures long-range dependencies from horizontal and narrow objects (e.g., the fifth to seventh rows of Figure~\ref{fig:seg}), while VRPM captures long-range correlations from vertical and narrow objects (e.g., the eighth to tenth rows of Figure~\ref{fig:seg}).

As shown in Figure~\ref{fig:rpm}, given the feature $f_s {\in} C {\times} H {\times} W$ produced by SPM, we first feed it into a convolution layer to reduce the number of the channels and obtain a new feature $f_s^1 {\in} C/4 {\times} H {\times} W$. 
Note that we use $C$ to represent the number of channels of $f_s$ for simplicity, which is different from the one used in Figure~\ref{fig:spm}. 
Then $f_s^1$ is separately fed into HRPM and VRPM to capture both horizontal and vertical long-range dependencies. 
Specifically, in HRPM, $f_s^1$ is first fed into a horizontal rectangle-shape pooling layer to obtain a new feature $f_h{\in} C/4 {\times} 1 {\times} W$. 
After that, we put $f_h$ through a 1D convolutional layer to obtain the feature $\hat{f_h}$.
Next, an upsampling operation is performed on $\hat{f_h}$ to expand the spatial size and then output the feature $\tilde{f_h}$.
Similarly, in VRPM, $f_s^1$ is first fed into a vertical rectangle-shape pooling layer to obtain a new feature $f_v{\in} C/4 {\times} H {\times} 1$.  After that, we use a 1D convolutional layer to obtain the feature $\hat{f_v}$.
Next, an upsampling operation is performed on $\hat{f_v}$ to expand the spatial size and then output the feature $\tilde{f_v}$.
We finally sum both $\tilde{f_h}$ and $\tilde{f_v}$ incorporating both horizontal and vertical long-range dependencies,
\begin{equation}
\begin{aligned}
\tilde{f_s} = \tilde{f_h} + \tilde{f_v}.
\end{aligned}
\end{equation}

Taking into account the advantages of short-range dependency modeling, we also consider incorporating SPM in our PRM to make the feature representations more discriminative.
As shown in Figure~\ref{fig:rpm}, we first feed $f_s$ into another convolution layer to reduce the number of the channels and obtain a new feature $f_s^2{\in} C/4 {\times} H {\times} W$.
We then capture short-range semantic dependencies by using Equation~\eqref{eq:local}.
\begin{equation}
\begin{aligned}
f_s^{'}={\rm conv}({\rm conv}(f_s^2) + {\rm SPM}(f_s^2)),
\end{aligned}
\label{eq:local}
\end{equation}
where ${\rm SPM}(\cdot)$ is the model proposed in Figure~\ref{fig:spm}.
After that, we combine both short-range and long-range semantic dependencies by using,
\begin{equation}
\begin{aligned}
f_r^{'} = {\rm conv}({\rm concat}(f_s^{'}, \tilde{f_s})).
\end{aligned}
\end{equation}
In this way, $f_r^{'}$ is more discriminative than $f_s$ by aggregating different types of contextual information via various pooling operations, leading to better results.
Finally, we also propose two methods to add the input feature $f_s$, constituting a residual connection \cite{he2016deep}.
The first one (RPM-I) is performing an element-wise addition:
\begin{equation}
\begin{aligned}
f_r = f_r^{'} + f_s.
\end{aligned}
\label{eq:add}
\end{equation}
The second one (RPM-II) is performing a channel-wise concatenation:
\begin{equation}
\begin{aligned}
f_r = {\rm concat}(f_r^{'}, f_s).
\end{aligned}
\label{eq:concat}
\end{equation}

\begin{figure}[t]
	\centering
	\subfigure[F-I]{\includegraphics[width=0.515\linewidth]{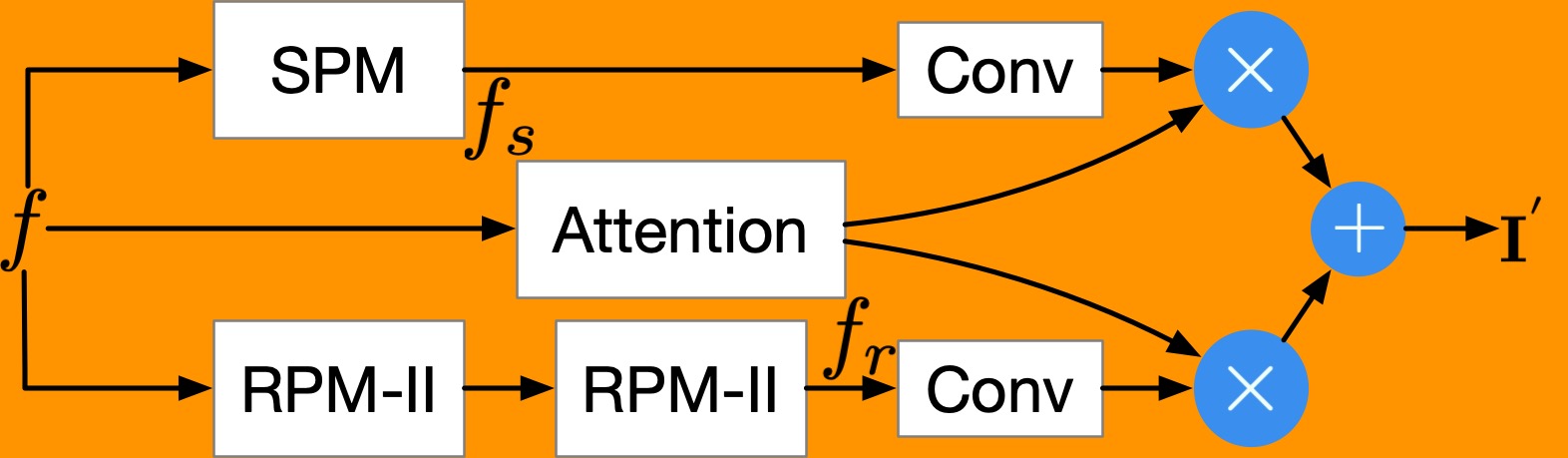}}
	\subfigure[F-II]{\includegraphics[width=0.47\linewidth]{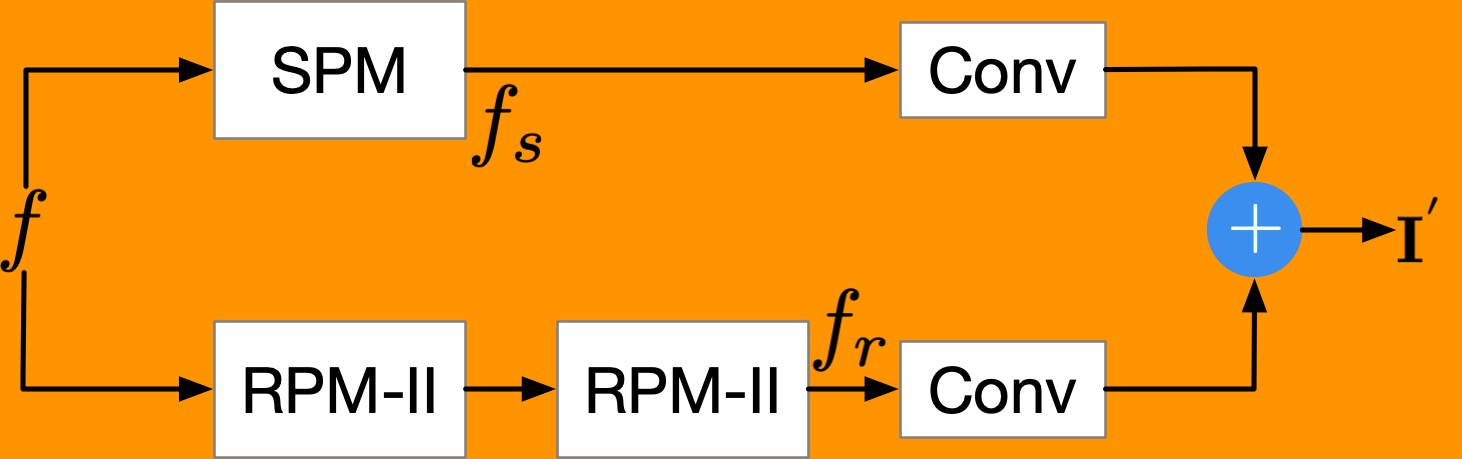}}
	\subfigure[F-III]{\includegraphics[width=0.32\linewidth]{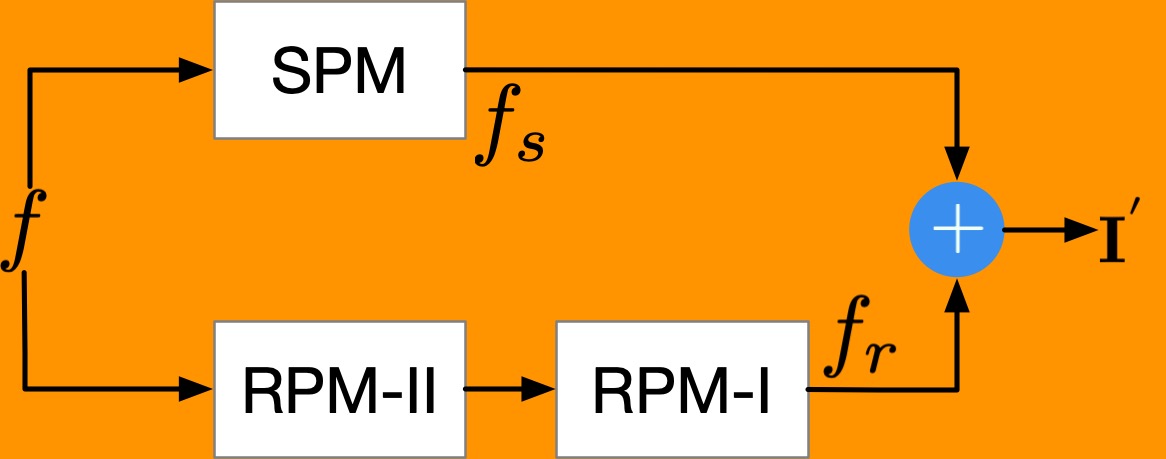}}
	\subfigure[F-IV]{\includegraphics[width=0.32\linewidth]{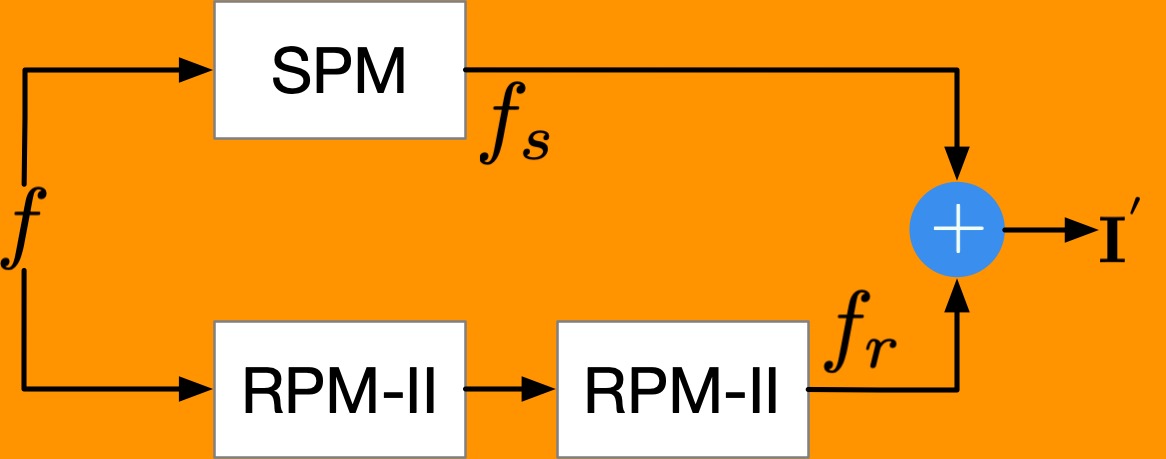}}
	\subfigure[F-V]{\includegraphics[width=0.32\linewidth]{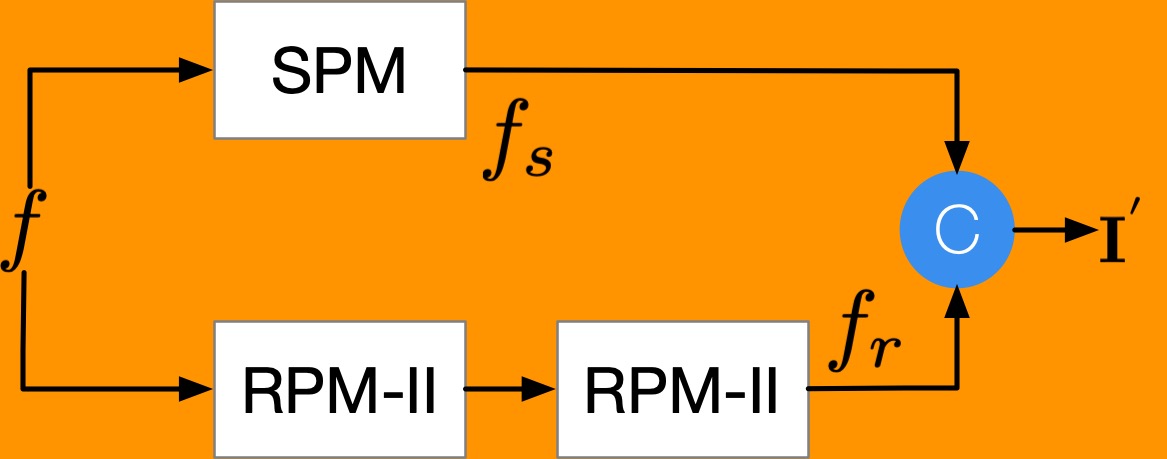}}
	\subfigure[F-VI]{\includegraphics[width=0.49\linewidth]{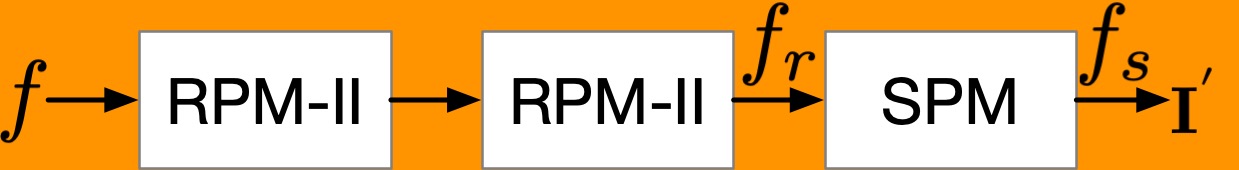}}
	\subfigure[F-VII]{\includegraphics[width=0.49\linewidth]{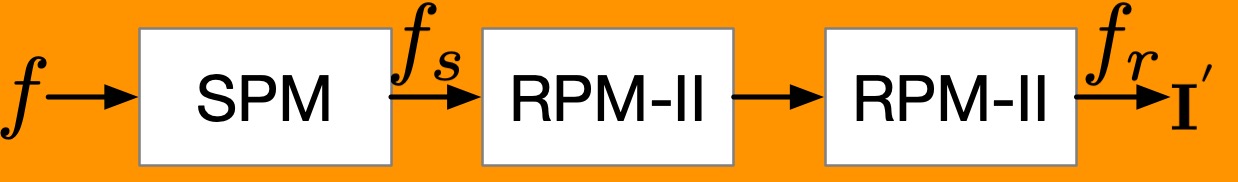}}
	\caption{The proposed seven image-level (a-b) and feature-level (c-g) fusion strategies. The symbols $\oplus$, $\otimes$, and $\textcircled{c}$ denote element-wise addition, element-wise multiplication, and channel-wise concatenation, respectively.}
	\label{fig:fusion}
\end{figure}

\begin{figure}[!t]
	\centering
	\includegraphics[width=1\linewidth]{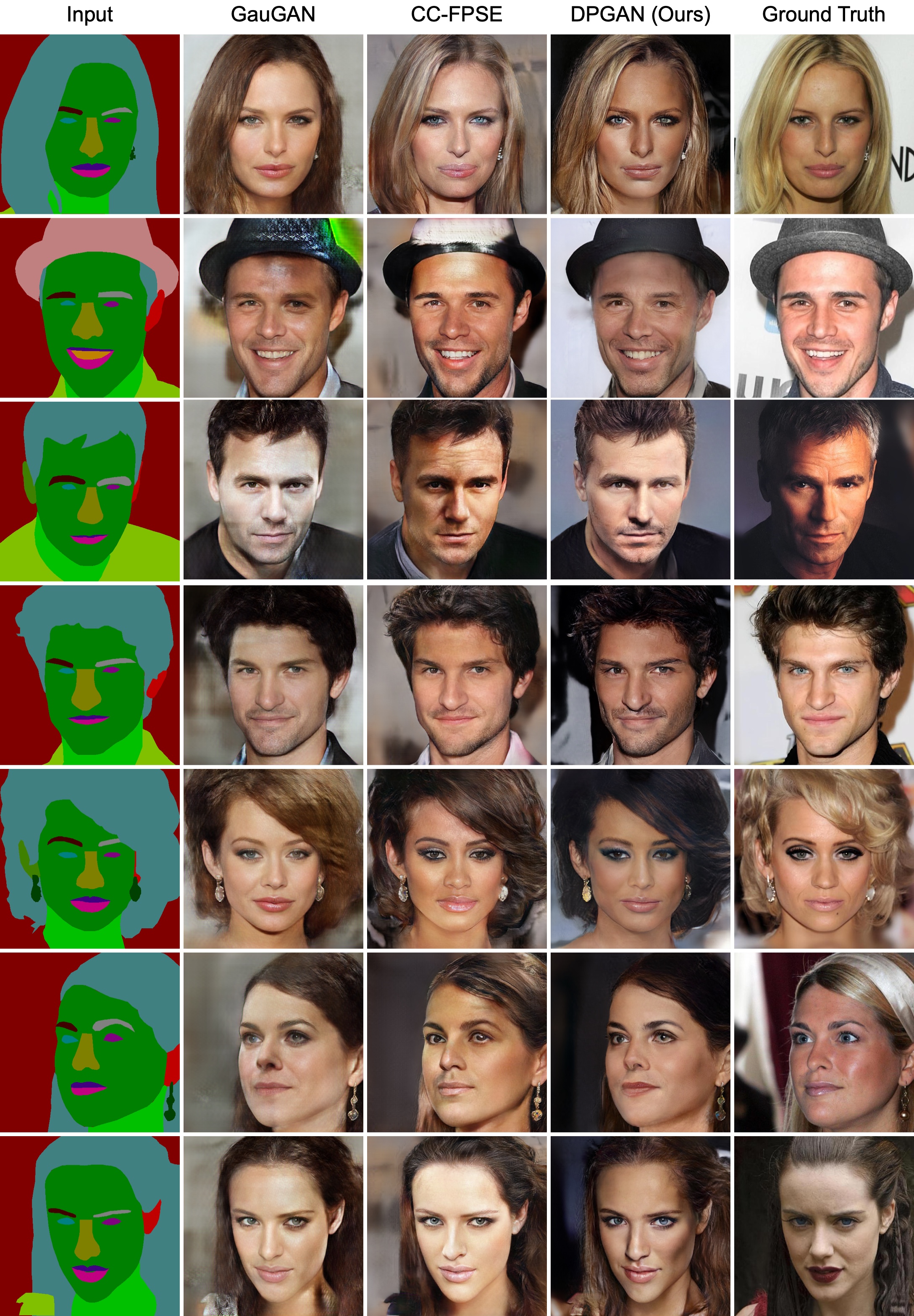}
	\caption{Qualitative comparison on CelebAMask-HQ. From left to right: Input, GauGAN~\cite{park2019semantic}, CC-FPSE~\cite{liu2019learning}, DPGAN (Ours), and Ground Truth. We see than DPGAN generates more convincing details than GauGAN and CC-FPSE, e.g., the hair, the hat, and the face skin in the first, second, and third row, respectively.}
	\label{fig:celeba_results}
\end{figure}

\noindent \textbf{Fusion of SPM and RPM.}
To take full advantage of both short-range and long-range semantic information, we further aggregate the outputs from these two pooling modules. 
Specifically, we propose seven aggregation methods as shown in Figure~\ref{fig:fusion}:
\begin{itemize}
\item (1) F-I is an image-level fusion strategy based on the attention fusion method proposed in LGGAN \cite{tang2020local}. The formulation of F-I can be expressed as: $\mathbf{I}^{'}{=} {\rm conv}(f_s) {\times} {A_1}  {+} {\rm conv}(f_r) {\times} A_2 $, where $A_1$ and $A_2$ are attention masks produced by an attention decoder.
\item (2) F-II is also an image-level fusion method represented as: $\mathbf{I}^{'}{=} \frac{{\rm conv}(f_s)  {+} {\rm conv}(f_r) }{2}$.
\item (3) The other five methods are feature-level based fusion: F-III, F-IV and F-V are parallel structures, while F-VI and F-VII are cascading structures. Mathematically, F-III can be written as $\mathbf{I}^{'}{=} {\rm conv}(f_s  {+} f_r)$.
\item (4) The difference between F-IV and F-III is that F-IV uses two `RPM-II' to produce the feature $f_r$, while F-III adopts one `RPM-II' and one `RPM-I' to generate $f_r$.
\item (5) The difference between F-V and F-IV is that F-V uses a channel-wise concatenation operation to combine both $f_s$ and $f_r$, thus it can be expressed as: $\mathbf{I}^{'}{=} {\rm conv} ({\rm concat}(f_s, f_r))$.
\item (6) Both F-VI and VII are cascading structures, and the difference between them is the order of RPM and SPM. We can represent F-VI as: $\mathbf{I}^{'}{=} {\rm conv} (f_s)$.
\item (7) F-VII is also shown in Figure~\ref{fig:method}, and it can be expressed as: $\mathbf{I}^{'}{=} {\rm conv} (f_r)$.
\end{itemize}

\begin{figure}[!t]
	\centering
	\includegraphics[width=1\linewidth]{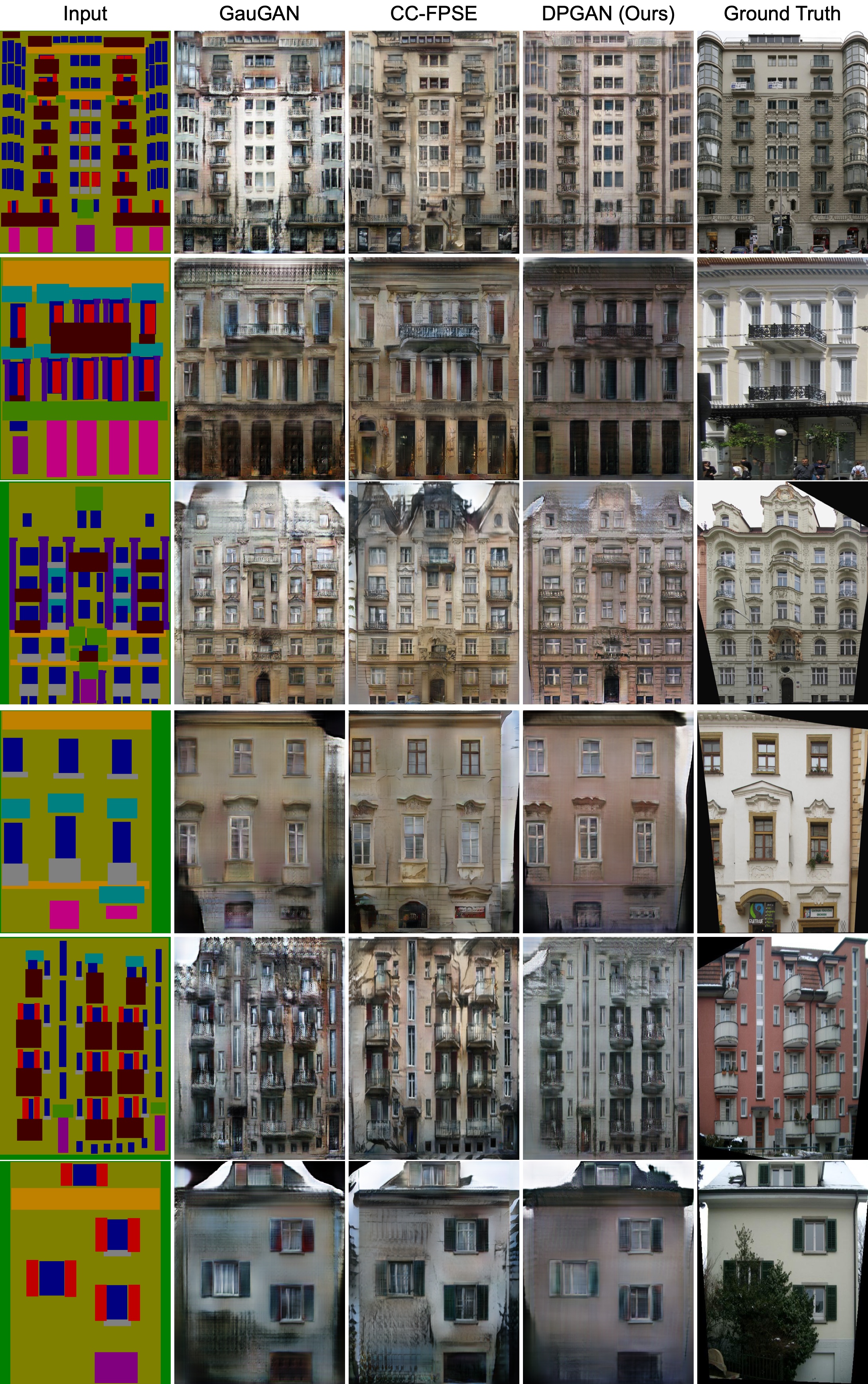}
	\caption{Qualitative comparison on Facades. From left to right: Input, GauGAN~\cite{park2019semantic}, CC-FPSE~\cite{liu2019learning}, DPGAN (Ours), and Ground Truth. We see that DPGAN generates more clear architecture structures with fewer artifacts than GauGAN and CC-FPSE.}
	\label{fig:facades_results}
\end{figure}

\noindent \textbf{Optimization Objective.}
We follow GauGAN \cite{park2019semantic} and employ three different losses as our optimization objective, i.e.,
$\mathcal{L} {=}  \lambda_{gan} \mathcal{L}_{gan} {+} \lambda_{f} \mathcal{L}_{f} {+} \lambda_{p} \mathcal{L}_{p}$,
where $\mathcal{L}_{gan}$, $\mathcal{L}_{f}$ and $\mathcal{L}_{p}$ denote adversarial loss, discriminator feature matching loss, and perceptual loss, respectively.
We set $\lambda_{gan}{=}1$, $\lambda_{f}{=}10$, and $\lambda_{p}{=}10$ in our experiments.

\noindent \textbf{Training Details.}
We use the multi-scale discriminator \cite{park2019semantic} as our discriminator $D$.
We use the Adam solver \cite{kingma2014adam} and set $\beta_1{=}0$, $\beta_2{=}0.999$.
Moreover, we set $n{=}4$ in the proposed SPM, and set $h_1{=}w_1{=}1$, $h_2{=}w_2{=}2, h_3{=}w_3{=}3$, and $h_4{=}w_4{=}6$, respectively.
The kernel size of convolutional layers in the proposed SPM is set to $1{\times}1$.
We set $n{=}2$ for SPM used in RPM, and set $h_1{=}w_1{=}12$ and $h_2{=}w_2{=}20$.
The kernel size of 1D convolutional layers in RPM is $1{\times}3$ and $3{\times}1$, respectively.
The proposed DPGAN is implemented by using PyTorch \cite{paszke2019pytorch}.
We conduct the experiments on NVIDIA DGX1 with 8 32GB V100 GPUs.

%% file: 4Experiments.tex
\section{Experiments}

\begin{figure}[!t] \small
	\centering
	\includegraphics[width=1\linewidth]{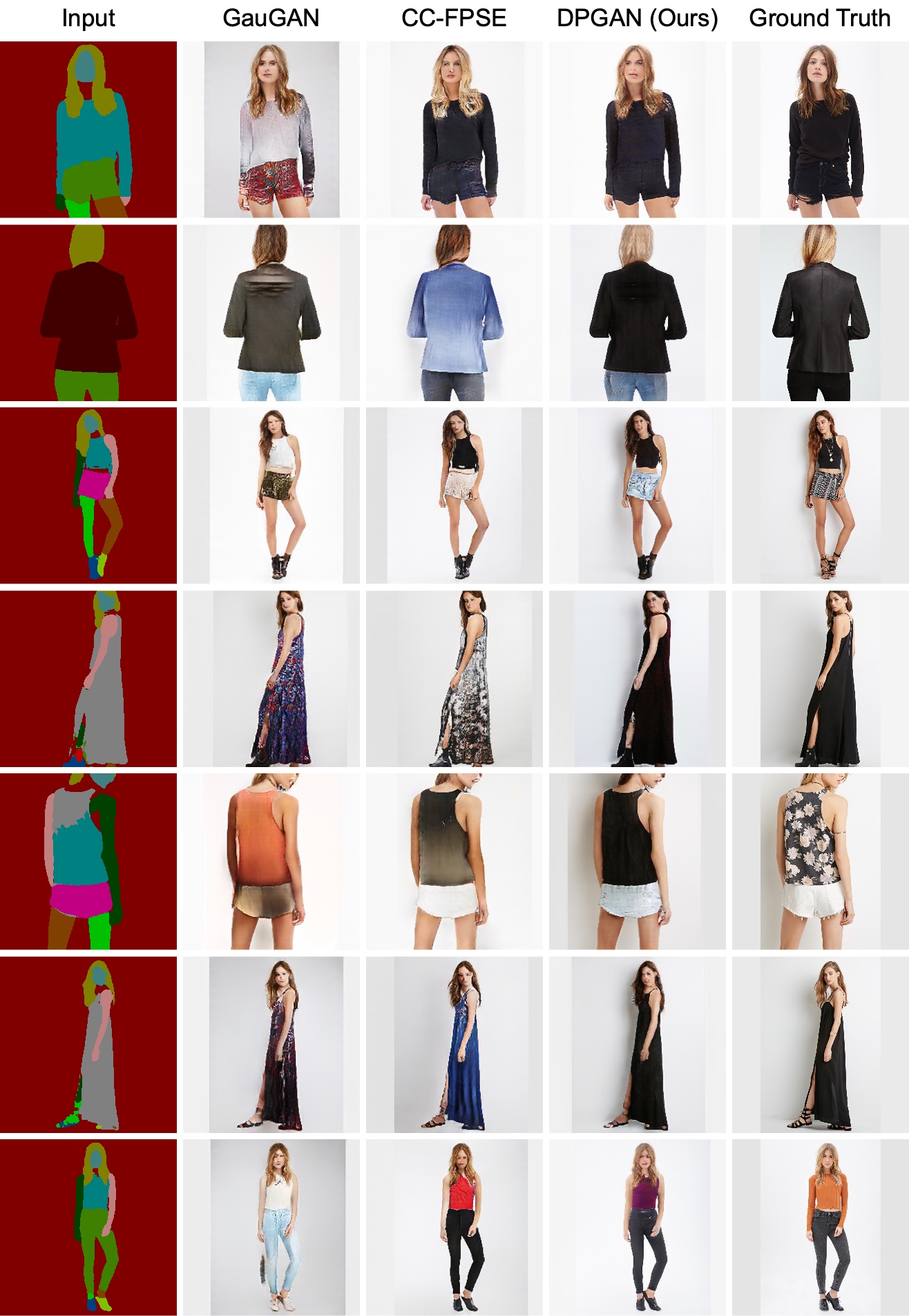}
	\caption{Qualitative comparison on DeepFashion. From left to right: Input, GauGAN~\cite{park2019semantic}, CC-FPSE~\cite{liu2019learning}, DPGAN (Ours), and Ground Truth. We see that DPGAN generates more photo-realistic clothes than GauGAN and CC-FPSE.}
	\label{fig:deepfashion_results}
\end{figure}

\begin{figure*}[t] \small
	\centering
	\includegraphics[width=1\linewidth]{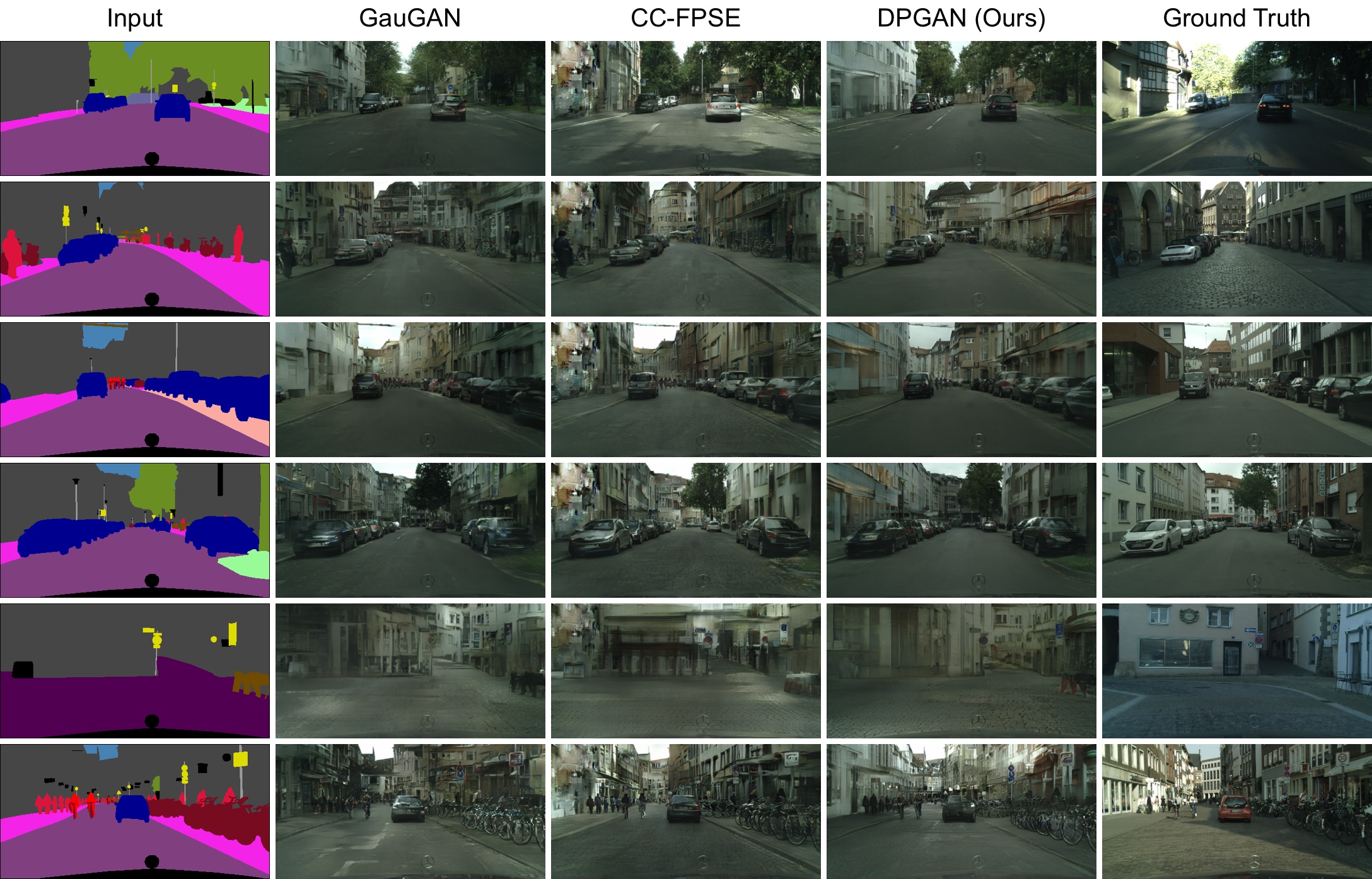}
	\caption{Qualitative comparison on Cityscapes. From left to right: Input, GauGAN~\cite{park2019semantic}, CC-FPSE~\cite{liu2019learning}, DPGAN (Ours), and Ground Truth. We see that DPGAN produces more realistic images with fewer artifacts than both leading methods.}
	\label{fig:city_results}
\end{figure*}

\begin{table*}[!t]
	\centering
		\caption{User study. The numbers indicate the percentage of users who favor the results of our proposed DPGAN over the competing methods.}
	\begin{tabular}{lccccc} \toprule
		AMT $\uparrow$                                           & Cityscapes & ADE20K & DeepFashion & Facades & CelebAMask-HQ \\ \midrule
		Ours vs. GauGAN~\cite{park2019semantic} &  65.78        &  68.72    & 66.85            & 67.54     & 69.91 \\ 
		Ours vs. CC-FPSE~\cite{liu2019learning} &  62.21         &  64.36   &  63.16                   & 64.54            & 67.18        \\ \bottomrule
	\end{tabular}
	\label{tab:amt}
\end{table*}

\begin{figure*}[!t] \small
	\centering
	\includegraphics[width=1\linewidth]{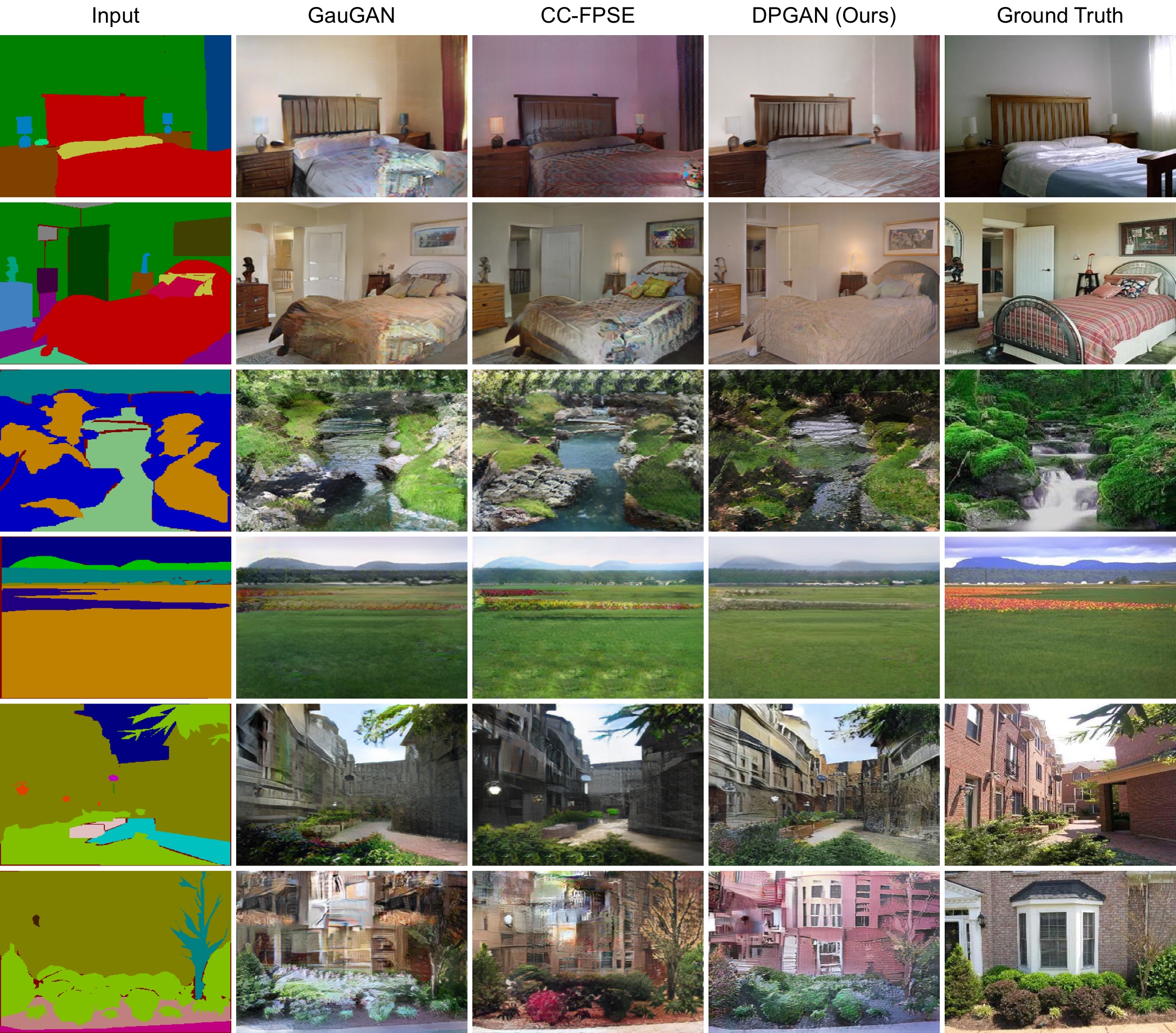}
	\caption{Qualitative comparison on ADE20K. From left to right: Input, GauGAN~\cite{park2019semantic}, CC-FPSE~\cite{liu2019learning}, DPGAN (Ours), and Ground Truth. We see that DPGAN produces realistic images while respecting the spatial semantic layout at the same time.}
	\label{fig:ade_results}
\end{figure*}

\noindent \textbf{Datasets.} We follow GauGAN~\cite{park2019semantic} and firstly conduct experiments on Cityscapes \cite{cordts2016cityscapes} and ADE20K \cite{zhou2017scene} datasets. Cityscapes contains street scene images, and ADE20K contains both indoor and outdoor scenes.
To further evaluate the robustness of our method, we conduct experiments on three more datasets with diverse scenarios, i.e., DeepFashion \cite{liu2016deepfashion}, CelebAMask-HQ \cite{CelebAMask-HQ}, and Facades \cite{tylevcek2013spatial}.
DeepFashion contains human body images, CelebAMask-HQ contains human facial images, and Facades contains facade images with diverse architectural styles.
Experiments are conducted using different image resolutions to validate that our DPGAN can also generate high-resolution images, i.e., ADE20K ($256 {\times} 256$), DeepFashion ($256 {\times} 256$), Cityscapes ($512 {\times} 256$), Facades ($512 {\times} 512$), and CelebAMask-HQ ($512 {\times} 512$). 

\noindent \textbf{Evaluation Metrics.}
We follow GauGAN~\cite{park2019semantic} and adopt mean Intersection-over-Union (mIoU), pixel accuracy (Acc), and Fr\'echet Inception Distance (FID)~\cite{heusel2017gans} as the evaluation metrics on Cityscapes and ADE20K.
For DeepFashion, CelebAMask-HQ, and Facades datasets, we use FID and Learned Perceptual Image Patch Similarity (LPIPS) \cite{zhang2018unreasonable} to evaluate the quality of the generated images.

\subsection{Comparisons with State-of-the-Art}
We adopt GauGAN \cite{park2019semantic} as our backbone and insert the proposed Double Pooling Module (DPM) before the last convolution layer to form our final model, i.e., DPGAN. 

\noindent \textbf{Qualitative Comparisons.}
We first compare the proposed DPGAN with GauGAN \cite{park2019semantic} and CC-FPSE~\cite{liu2019learning} on DeepFashion, CelebAMask-HQ, and Facades datasets.
Note that we used the source code provided by the authors to generate the results of GauGAN and CC-FPSE on these three datasets for fair comparisons.
Visualization results are shown in Figures~\ref{fig:celeba_results}, \ref{fig:facades_results}, and \ref{fig:deepfashion_results}.
We can see that the proposed DPGAN generates more photo-realistic and semantically-consistent results than both GauGAN and CC-FPSE.

Moreover, we compare DPGAN with two leading methods on both Cityscapes and ADE20K datasets, i.e., GauGAN \cite{park2019semantic} and CC-FPSE \cite{liu2019learning}.
Comparison results are shown in Figures~\ref{fig:city_results} and \ref{fig:ade_results}.
We can see that our DPGAN produces more clear and visually plausible results than both leading methods, further validating the effectiveness of our proposed DPGAN.

\noindent \textbf{User Study.}
We follow the same evaluation protocol of GauGAN and also perform a user study.
The results compared with GauGAN and CC-FPSE are shown in Table~\ref{tab:amt}.
We see that users strongly favor the results generated by our proposed DPGAN on all datasets, further validating that the generated images by our DPGAN are more photo-realistic.

\begin{table*}[!t]
	\centering
		\caption{Quantitative comparison of different methods on DeepFashion, Facades, and CelebAMask-HQ.}
	\begin{tabular}{lcccccc} \toprule
		\multirow{2}{*}{Method}  &  \multicolumn{2}{c}{DeepFashion} & \multicolumn{2}{c}{Facades} & \multicolumn{2}{c}{CelebAMask-HQ}\\ \cmidrule(lr){2-3} \cmidrule(lr){4-5} \cmidrule(lr){6-7} 
		& FID $\downarrow$ & LPIPS $\downarrow$  & FID $\downarrow$ & LPIPS $\downarrow$  & FID $\downarrow$ & LPIPS $\downarrow$ \\ \midrule
		GauGAN & 22.8 & 0.2476 & 116.8 & \textbf{0.5437} & 42.2 & 0.4870 \\
		+ DPM (Ours) & \textbf{20.8} & \textbf{0.2455} & \textbf{115.1} & 0.5503 & \textbf{25.1} & \textbf{0.4823}  \\
		\bottomrule
	\end{tabular}
	\label{tab:sota2}
\end{table*}

\begin{table*}[!t]
	\centering
		\caption{Quantitative comparison of different methods on Cityscapes and ADE20K. The results of other methods are reported from their papers.}
	\begin{tabular}{lllllll} \toprule
		\multirow{2}{*}{Method}  & \multicolumn{3}{c}{Cityscapes} & \multicolumn{3}{c}{ADE20K} \\ \cmidrule(lr){2-4} \cmidrule(lr){5-7} 
		& mIoU $\uparrow$ & Acc $\uparrow$  & FID  $\downarrow$ & mIoU $\uparrow$    & Acc $\uparrow$  & FID  $\downarrow$  \\ \midrule
		CRN~\cite{chen2017photographic} & 52.4  & 77.1 & 104.7  & 22.4 & 68.8 & 73.3 \\
		SIMS~\cite{qi2018semi}  & 47.2  & 75.5 & 49.7 & -  & - & -  \\
		Pix2pixHD~\cite{wang2018high}    & 58.3  & 81.4 & 95.0  & 20.3 & 69.2 & 81.8  \\ 
		GAN Compression~\cite{li2020gan} & 61.2 & - & - & - & - & - \\
		BachGAN \cite{li2020bachgan} & - & 70.4 & 73.3 & - & 66.8 & 49.8 \\
		PIS \cite{dundar2020panoptic} & 64.8 & 82.4 & 96.4 & - & - & - \\ 
		SelectionGAN~\cite{tang2019multi} & 63.8 & 82.4 & 65.2 & 40.1 & 81.2 & 33.1\\
		DAGAN~\cite{tang2020dual} & 66.1 & 82.6 & 60.3 & 40.5 & 81.6 & 31.9 \\
		LGGAN~\cite{tang2020local} & 68.4 & 83.0 & 57.7 & 41.6 & 81.8 & 31.6 \\ \hline
		GauGAN~\cite{park2019semantic} & 62.3  & 81.9 & 71.8  & 38.5 & 79.9 & 33.9  \\
		+ DPM (Ours) & 65.2 (\textbf{+2.9}) & 82.6 (\textbf{+0.7}) & 53.0 (\textbf{-18.8}) & 39.2 (\textbf{+0.7}) & 80.4 (\textbf{+0.5}) & 31.7 (\textbf{-2.2})  \\ \hline
		CC-FPSE~\cite{liu2019learning}  & 65.5 & 82.3 & 54.3 & 43.7 & 82.9 & 31.7  \\ 
		+ DPM (Ours) & 66.9 (\textbf{+1.4}) & 82.8 (\textbf{+0.5}) & 51.9 (\textbf{-2.4}) & 44.8 (\textbf{+1.1}) & 83.2 (\textbf{+0.3}) & 30.3 (\textbf{-1.4}) \\ \hline
		TSIT \cite{jiang2020tsit} & 65.9  & 82.7 & 59.2 & 38.6 & 80.8 & 31.6  \\
		+ DPM (Ours) & 66.7 (\textbf{+0.8}) & 83.1 (\textbf{+0.4}) & 56.1 (\textbf{-3.1}) & 39.9 (\textbf{+1.3}) & 81.2 (\textbf{+0.4}) & 30.5 (\textbf{-1.1}) \\ 
		\bottomrule
	\end{tabular}
	\label{tab:sota1}
\end{table*}

\noindent \textbf{Quantitative Comparisons.}
Although the user study is more suitable for evaluating the quality of the generated image in this task, we also follow GauGAN and use mIoU, Acc, FID, and LPIPS for quantitative evaluation.
The results compared with several leading methods are shown in Tables \ref{tab:sota2} and \ref{tab:sota1}.
Firstly, we observe that the proposed DPGAN achieves the best results compared with GauGAN on DeepFashion, CelebAMask-HQ, and Facades datasets, as shown in Table~\ref{tab:sota2}.
Moreover, we can see that DPGAN achieves competitive results compared with other leading methods on both Cityscapes and ADE20K datasets in Table \ref{tab:sota1}.

\noindent \textbf{Visualization of Generated Semantic Maps.}
We follow GauGAN and adopt the pretrained DRN-D-105 \cite{yu2017dilated} on the generated Cityscapes images to produce semantic maps.
The results compared with those produced by GauGAN are shown in Figure~\ref{fig:seg}.
We clearly see that the proposed SPM and RPM can capture short-range and long-range semantic dependencies, leading to more semantically-consistent and realistic results than GauGAN.

\subsection{Ablation Study}

\noindent \textbf{Baselines of DPGAN.} 
We conduct an extensive ablation study on Cityscapes to evaluate each component of the proposed DPGAN. 
DPGAN has 13 baselines (i.e., B1-B13) as shown in Table \ref{tab:abla}. 
\begin{itemize}
\item (1) B1 is our baseline and uses a GauGAN structure~\cite{park2019semantic}.
\item (2) B2 uses the proposed Square-shape Pooling Module (SPM) to capture short-range semantic dependencies. 
\item (3) B3 employs the proposed Rectangle-shape Pooling Module (RPM) to capture long-range semantic dependencies from both horizontal and vertical directions.
Note that B3 uses Equation~\eqref{eq:add} to generate the feature $f_r$. 
\item (4) The difference between B4 and B3 is that B4 uses Equation~\eqref{eq:concat} to generate $f_r$. 
\item (5) B5 is based on B3 and uses the proposed RPM twice.
\item (6) B6 is based on B4 and uses the proposed RPM twice.
\item (7)-(13) B7 to B13 are seven fusion methods proposed in Figure~\ref{fig:fusion}, which aim to effectively combine both short-range and long-range dependencies for further enlarging the receptive field of our model. 
\end{itemize}

\begin{table}[!t]
	\centering
		\caption{Ablation study of our DPGAN on Cityscapes.}
	\begin{tabular}{clccc} \toprule
		No. & Setting          &  mIoU $\uparrow$ & Acc $\uparrow$ & FID $\downarrow$  \\ \midrule	
		B1 & GauGAN~\cite{park2019semantic} & 62.3 & 81.9 & 71.8 \\ \hline
		B2 & B1 + SPM            & 64.9 & 82.5 & 55.9 \\ \hline
		B3 & B1 + RPM-I          & 63.9 & 82.3 & 55.4  \\
		B4 & B1 + RPM-II         & 63.8 & 82.4 & 54.9  \\ 
		B5 & B1 + 2 RPM-I        & 64.5 & 82.4 & 54.2  \\
		B6 & B1 + 2 RPM-II       & 64.7 & 82.5 & 53.2  \\  \hline
		B7 & B6 + SPM + F-I      & 64.2 & 82.5 & 53.2\\ 
		B8 & B6 + SPM + F-II     & \textbf{65.2} & 82.5 & 54.5 \\ 
		B9 & B6 + SPM + F-III    & 64.8 & \textbf{82.6} & 53.3\\ 
		B10 & B6 + SPM + F-IV    & 65.0 & \textbf{82.6} & 53.6\\  
		B11 & B6 + SPM + F-V     & 63.1 & 82.4 & 53.3 \\  
		B12 & B6 + SPM + F-VI    & 64.8 & 82.4 & 53.4 \\  
		B13 & B6 + SPM + F-VII   & \textbf{65.2} & \textbf{82.6} & \textbf{53.0} \\  \bottomrule
	\end{tabular}
	\label{tab:abla}
\end{table}

\noindent \textbf{Ablation Analysis.}
The results of the ablation study are shown in Table \ref{tab:abla}.
We can see that B2 achieves better results than B1 on all metrics, which confirms the importance of modeling short-range semantic dependencies.
Both B3 and B4 outperform B1, confirming the effectiveness of modeling long-range semantic dependencies.
B6 outperforms B4, and B5 outperforms B3, showing that adding more layers of RPM will further enlarge the receptive field.
Moreover, we observe that B13 outperforms B6 on all metrics, showing that both short-range and long-range semantic dependencies are essential for generating high-quality results.
Lastly, we observe that B13 achieves better results than B7-B12, demonstrating the effectiveness of the fusion strategy F-VII.

We also note that B7 and B11 achieve slightly deteriorated results compared with B6 on the mIoU metric, which means both F-I and F-V are not very suitable fusion strategies for the proposed SPM and RPM. Other fusion strategies such as F-II, F-IV, and F-VII
achieve slightly different results in all the evaluation metrics.
This is because the proposed SPM and RPM are powerful to capture both short-term and long-term dependencies and a simple fusion strategy such as F-II, F-IV, or F-VII can achieve good generation performance. This also further illustrates the effectiveness of both SPM and RPM.

\begin{figure}[t!]
	\centering
	\includegraphics[width=1\linewidth]{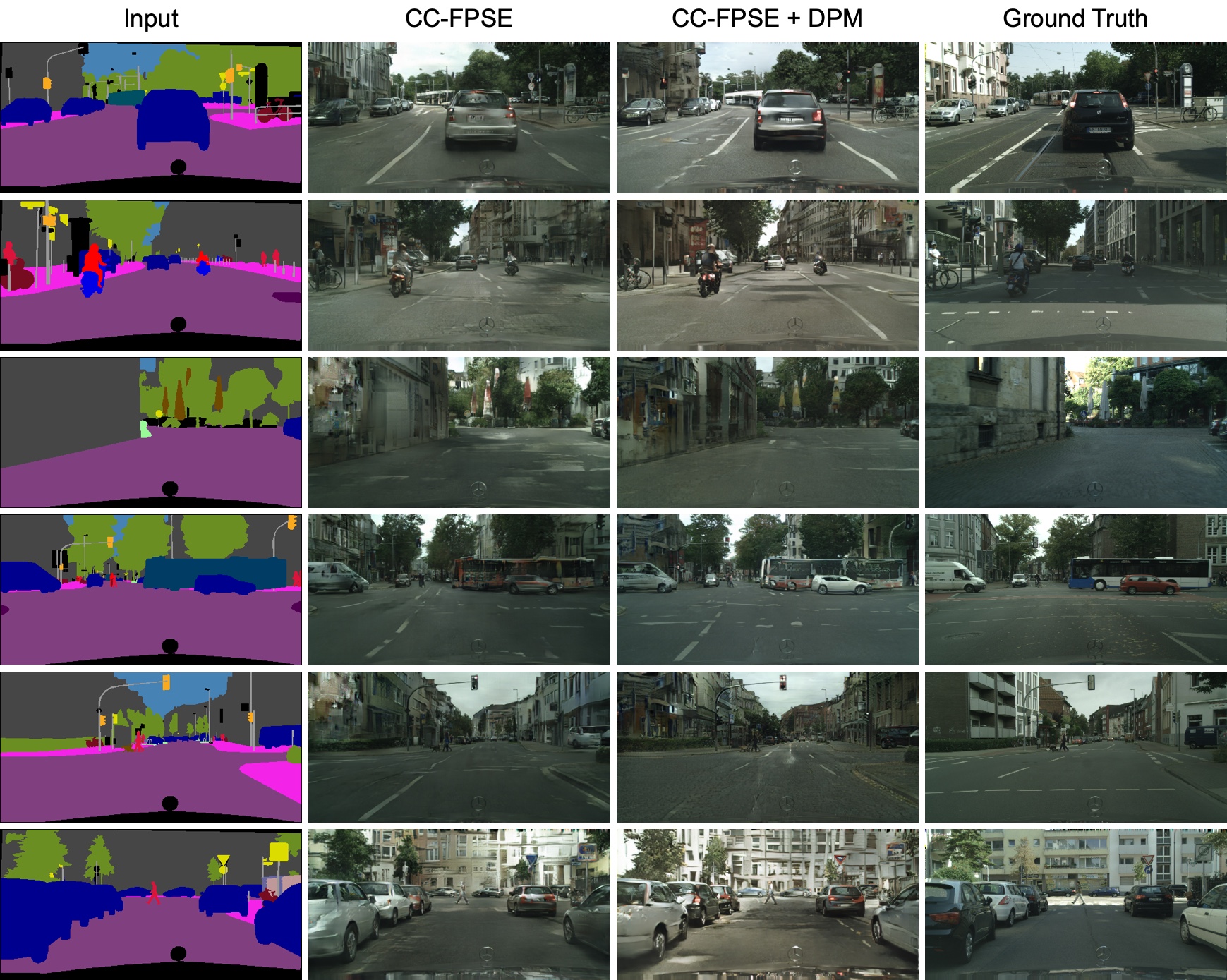}
	\caption{The generalization ability of the proposed DPM on Cityscapes. We see that the CC-FPSE model with DPM (CC-FPSE + DPM) generates more realistic images with fewer artifacts than the CC-FPSE model without using DPM.}
	\label{fig:results_city_supp}
\end{figure}

\begin{figure}[t!]
	\centering
	\includegraphics[width=1\linewidth]{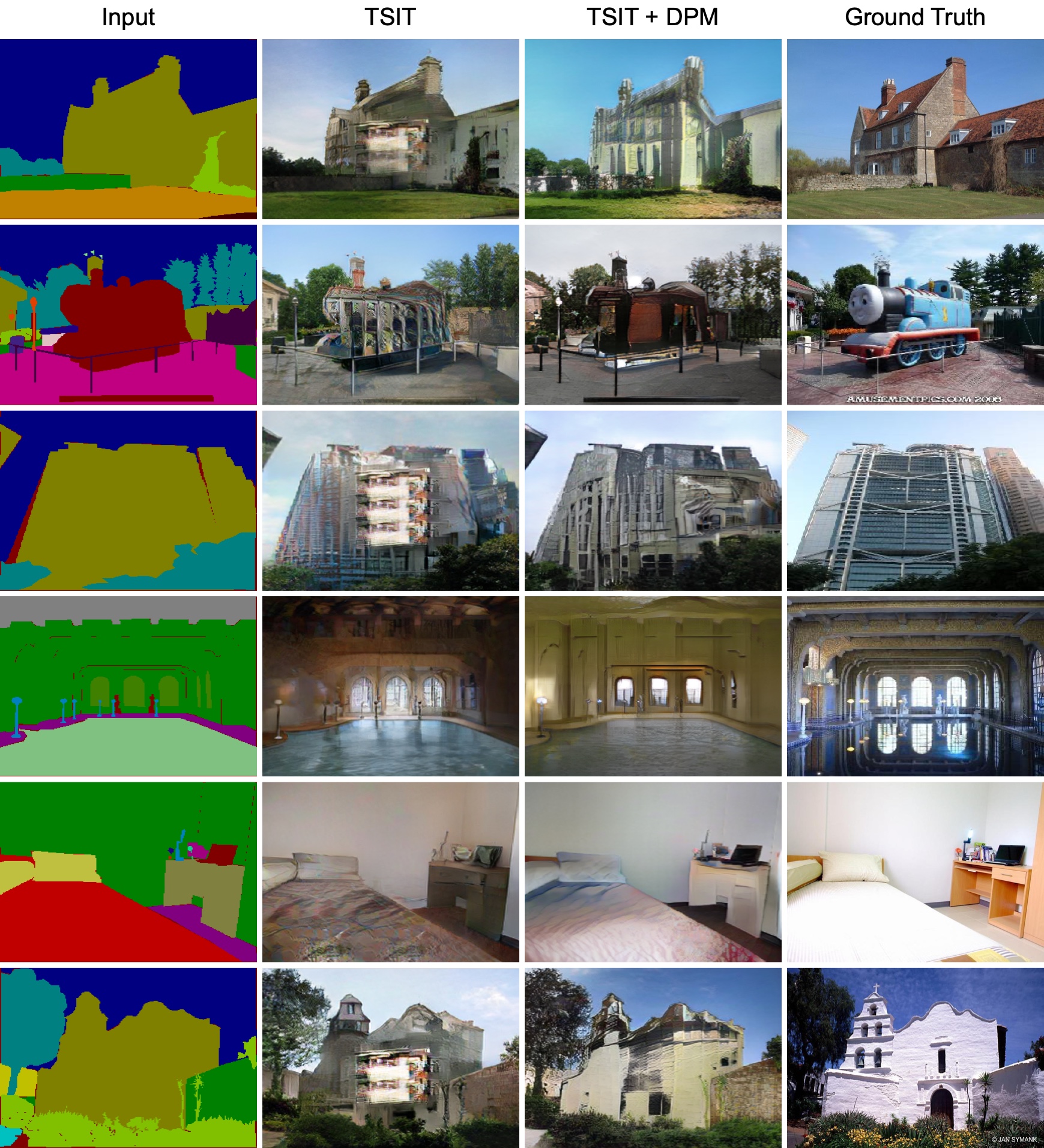}
	\caption{The generalization ability of the proposed DPM on ADE20K. We see that the TSIT model with DPM (TSIT + DPM) generates more realistic images with fewer artifacts than the TSIT model without using DPM.}
	\label{fig:ade_results_supp}
\end{figure}

\noindent \textbf{Generalization of DPM.} 
The proposed Double Pooling Module (DPM) is general and can be seamlessly integrated into any existing GAN-based architecture to improve the image translation performance.
Therefore, to validate the generalization ability of the proposed DPM, we further conduct more experiments on both Cityscapes and ADE20K datasets.
Specifically, we adopt CC-FPSE as our $E$ and then combine CC-FPSE and our DPM to form the final model.
We observe that the CC-FPSE model with our DPM (i.e., CC-FPSE + DPM) further improves all the three metrics, as shown in Table~\ref{tab:sota1}.
In the visualization results shown in Figure~\ref{fig:results_city_supp}, we clearly observe that the CC-FPSE model with our DPM generates more realistic images with fewer artifacts than the CC-FPSE model without using our DPM on Cityscapes.

Moreover, we employ TSIT as our $E$ and then combine TSIT and our DPM to form the final model.
We observe that the TSIT model with our DPM (i.e., TSIT + DPM) further improves all the three metrics compared with the original TSIT, as shown in Table~\ref{tab:sota1}.
We also provide the visualization results in Figure~\ref{fig:ade_results_supp}, we clearly observe that the TSIT model with our DPM generates more realistic images with fewer artifacts than the TSIT model without using our DPM on ADE20K. Both experimental results validate that the proposed DPM can be integrated with other methods to further boost the translation performance.

%% file: 5Conclusion.tex
\section{Conclusions}
We propose a novel Double Pooling GAN (DPGAN) for the challenging layout-to-image translation task.
Specifically, we present a novel Double Pooling Module, which consists of the Square-shape Pooling Module (SPM) and the Rectangle-shape Pooling Module (RPM). 
SPM is used to capture short-range and local semantic dependencies. 
RPM is used to capture long-range and global semantic dependencies from both horizontal and vertical directions.
The outputs of SPM and RPM are combined with the proposed fusion strategies to further effectively enlarge the receptive field of our model. 
Extensive experiments on five popular datasets demonstrate that DPGAN establishes new state-of-the-art results.

%% file: tip.bbl
\begin{thebibliography}{10}
\providecommand{\url}[1]{#1}
\csname url@samestyle\endcsname
\providecommand{\newblock}{\relax}
\providecommand{\bibinfo}[2]{#2}
\providecommand{\BIBentrySTDinterwordspacing}{\spaceskip=0pt\relax}
\providecommand{\BIBentryALTinterwordstretchfactor}{4}
\providecommand{\BIBentryALTinterwordspacing}{\spaceskip=\fontdimen2\font plus
\BIBentryALTinterwordstretchfactor\fontdimen3\font minus
  \fontdimen4\font\relax}
\providecommand{\BIBforeignlanguage}[2]{{%
\expandafter\ifx\csname l@#1\endcsname\relax
\typeout{** WARNING: IEEEtran.bst: No hyphenation pattern has been}%
\typeout{** loaded for the language `#1'. Using the pattern for}%
\typeout{** the default language instead.}%
\else
\language=\csname l@#1\endcsname
\fi
#2}}
\providecommand{\BIBdecl}{\relax}
\BIBdecl

\bibitem{goodfellow2014generative}
I.~Goodfellow, J.~Pouget-Abadie, M.~Mirza, B.~Xu, D.~Warde-Farley, S.~Ozair,
  A.~Courville, and Y.~Bengio, ``Generative adversarial nets,'' in
  \emph{NeurIPS}, 2014.

\bibitem{chen2017photographic}
Q.~Chen and V.~Koltun, ``Photographic image synthesis with cascaded refinement
  networks,'' in \emph{ICCV}, 2017.

\bibitem{isola2017image}
P.~Isola, J.-Y. Zhu, T.~Zhou, and A.~A. Efros, ``Image-to-image translation
  with conditional adversarial networks,'' in \emph{CVPR}, 2017.

\bibitem{wang2018high}
T.-C. Wang, M.-Y. Liu, J.-Y. Zhu, A.~Tao, J.~Kautz, and B.~Catanzaro,
  ``High-resolution image synthesis and semantic manipulation with conditional
  gans,'' in \emph{CVPR}, 2018.

\bibitem{park2019semantic}
T.~Park, M.-Y. Liu, T.-C. Wang, and J.-Y. Zhu, ``Semantic image synthesis with
  spatially-adaptive normalization,'' in \emph{CVPR}, 2019.

\bibitem{liu2019learning}
X.~Liu, G.~Yin, J.~Shao, X.~Wang \emph{et~al.}, ``Learning to predict
  layout-to-image conditional convolutions for semantic image synthesis,'' in
  \emph{NeurIPS}, 2019.

\bibitem{dundar2020panoptic}
A.~Dundar, K.~Sapra, G.~Liu, A.~Tao, and B.~Catanzaro, ``Panoptic-based image
  synthesis,'' in \emph{CVPR}, 2020.

\bibitem{jiang2020tsit}
L.~Jiang, C.~Zhang, M.~Huang, C.~Liu, J.~Shi, and C.~C. Loy, ``Tsit: A simple
  and versatile framework for image-to-image translation,'' in \emph{ECCV},
  2020.

\bibitem{tang2020local}
H.~Tang, D.~Xu, Y.~Yan, P.~H. Torr, and N.~Sebe, ``Local class-specific and
  global image-level generative adversarial networks for semantic-guided scene
  generation,'' in \emph{CVPR}, 2020.

\bibitem{zhou2017scene}
B.~Zhou, H.~Zhao, X.~Puig, S.~Fidler, A.~Barriuso, and A.~Torralba, ``Scene
  parsing through ade20k dataset,'' in \emph{CVPR}, 2017.

\bibitem{liu2016deepfashion}
Z.~Liu, P.~Luo, S.~Qiu, X.~Wang, and X.~Tang, ``Deepfashion: Powering robust
  clothes recognition and retrieval with rich annotations,'' in \emph{CVPR},
  2016.

\bibitem{cordts2016cityscapes}
M.~Cordts, M.~Omran, S.~Ramos, T.~Rehfeld, M.~Enzweiler, R.~Benenson,
  U.~Franke, S.~Roth, and B.~Schiele, ``The cityscapes dataset for semantic
  urban scene understanding,'' in \emph{CVPR}, 2016.

\bibitem{CelebAMask-HQ}
C.-H. Lee, Z.~Liu, L.~Wu, and P.~Luo, ``Maskgan: Towards diverse and
  interactive facial image manipulation,'' in \emph{CVPR}, 2020.

\bibitem{tylevcek2013spatial}
R.~Tyle{\v{c}}ek and R.~{\v{S}}{\'a}ra, ``Spatial pattern templates for
  recognition of objects with regular structure,'' in \emph{GCPR}, 2013.

\bibitem{zhang2020dual}
J.~Zhang, J.~Chen, H.~Tang, W.~Wang, Y.~Yan, E.~Sangineto, and N.~Sebe, ``Dual
  in-painting model for unsupervised gaze correction and animation in the
  wild,'' in \emph{ACM MM}, 2020.

\bibitem{karras2019style}
T.~Karras, S.~Laine, and T.~Aila, ``A style-based generator architecture for
  generative adversarial networks,'' in \emph{CVPR}, 2019.

\bibitem{tang2020unified}
H.~Tang, H.~Liu, and N.~Sebe, ``Unified generative adversarial networks for
  controllable image-to-image translation,'' \emph{IEEE TIP}, vol.~29, pp.
  8916--8929, 2020.

\bibitem{liu2020exocentric}
G.~Liu, H.~Tang, H.~Latapie, and Y.~Yan, ``Exocentric to egocentric image
  generation via parallel generative adversarial network,'' in \emph{ICASSP},
  2020.

\bibitem{tang2021total}
H.~Tang and N.~Sebe, ``Total generate: Cycle in cycle generative adversarial
  networks for generating human faces, hands, bodies, and natural scenes,''
  \emph{IEEE TMM}, 2021.

\bibitem{liu2021cross}
G.~Liu, H.~Tang, H.~Latapie, J.~Corso, and Y.~Yan, ``Cross-view exocentric to
  egocentric video synthesis,'' in \emph{ACM MM}, 2021.

\bibitem{chen2021unsupervised}
H.~Chen, H.~Tang, H.~Shi, W.~Peng, N.~Sebe, and G.~Zhao, ``Intrinsic-extrinsic
  preserved gans for unsupervised 3d pose transfer,'' in \emph{ICCV}, 2021.

\bibitem{mirza2014conditional}
M.~Mirza and S.~Osindero, ``Conditional generative adversarial nets,''
  \emph{arXiv preprint arXiv:1411.1784}, 2014.

\bibitem{choi2018stargan}
Y.~Choi, M.~Choi, M.~Kim, J.-W. Ha, S.~Kim, and J.~Choo, ``Stargan: Unified
  generative adversarial networks for multi-domain image-to-image
  translation,'' in \emph{CVPR}, 2018.

\bibitem{tang2019expression}
H.~Tang, W.~Wang, S.~Wu, X.~Chen, D.~Xu, N.~Sebe, and Y.~Yan, ``Expression
  conditional gan for facial expression-to-expression translation,'' in
  \emph{ICIP}, 2019.

\bibitem{tang2019attribute}
H.~Tang, X.~Chen, W.~Wang, D.~Xu, J.~J. Corso, N.~Sebe, and Y.~Yan,
  ``Attribute-guided sketch generation,'' in \emph{FG 2019}, 2019.

\bibitem{zhang2017stackgan}
H.~Zhang, T.~Xu, H.~Li, S.~Zhang, X.~Wang, X.~Huang, and D.~N. Metaxas,
  ``Stackgan: Text to photo-realistic image synthesis with stacked generative
  adversarial networks,'' in \emph{ICCV}, 2017.

\bibitem{tao2020df}
M.~Tao, H.~Tang, S.~Wu, N.~Sebe, X.-Y. Jing, F.~Wu, and B.~Bao, ``Df-gan: Deep
  fusion generative adversarial networks for text-to-image synthesis,''
  \emph{arXiv preprint arXiv:2008.05865}, 2020.

\bibitem{tang2020xinggan}
H.~Tang, S.~Bai, L.~Zhang, P.~H. Torr, and N.~Sebe, ``Xinggan for person image
  generation,'' in \emph{ECCV}, 2020.

\bibitem{tang2019cycle}
H.~Tang, D.~Xu, G.~Liu, W.~Wang, N.~Sebe, and Y.~Yan, ``Cycle in cycle
  generative adversarial networks for keypoint-guided image generation,'' in
  \emph{ACM MM}, 2019.

\bibitem{chan2019everybody}
C.~Chan, S.~Ginosar, T.~Zhou, and A.~A. Efros, ``Everybody dance now,'' in
  \emph{ICCV}, 2019.

\bibitem{tang2020bipartite}
H.~Tang, S.~Bai, P.~H. Torr, and N.~Sebe, ``Bipartite graph reasoning gans for
  person image generation,'' in \emph{BMVC}, 2020.

\bibitem{tang2018gesturegan}
H.~Tang, W.~Wang, D.~Xu, Y.~Yan, and N.~Sebe, ``Gesturegan for hand
  gesture-to-gesture translation in the wild,'' in \emph{ACM MM}, 2018.

\bibitem{tang2019attention}
H.~Tang, D.~Xu, N.~Sebe, and Y.~Yan, ``Attention-guided generative adversarial
  networks for unsupervised image-to-image translation,'' in \emph{IJCNN},
  2019.

\bibitem{tang2019multi}
H.~Tang, D.~Xu, N.~Sebe, Y.~Wang, J.~J. Corso, and Y.~Yan, ``Multi-channel
  attention selection gan with cascaded semantic guidance for cross-view image
  translation,'' in \emph{CVPR}, 2019.

\bibitem{duan2021cascade}
B.~Duan, W.~Wang, H.~Tang, H.~Latapie, and Y.~Yan, ``Cascade attention guided
  residue learning gan for cross-modal translation,'' in \emph{ICPR}, 2021.

\bibitem{tang2021attentiongan}
H.~Tang, H.~Liu, D.~Xu, P.~H. Torr, and N.~Sebe, ``Attentiongan: Unpaired
  image-to-image translation using attention-guided generative adversarial
  networks,'' \emph{IEEE TNNLS}, 2021.

\bibitem{zhu2020sean}
P.~Zhu, R.~Abdal, Y.~Qin, and P.~Wonka, ``Sean: Image synthesis with semantic
  region-adaptive normalization,'' in \emph{CVPR}, 2020.

\bibitem{ntavelis2020sesame}
E.~Ntavelis, A.~Romero, I.~Kastanis, L.~Van~Gool, and R.~Timofte, ``Sesame:
  Semantic editing of scenes by adding, manipulating or erasing objects,'' in
  \emph{ECCV}, 2020.

\bibitem{zhu2020semantically}
Z.~Zhu, Z.~Xu, A.~You, and X.~Bai, ``Semantically multi-modal image
  synthesis,'' in \emph{CVPR}, 2020.

\bibitem{tang2020dual}
H.~Tang, S.~Bai, and N.~Sebe, ``Dual attention gans for semantic image
  synthesis,'' in \emph{ACM MM}, 2020.

\bibitem{tang2020edge}
H.~Tang, X.~Qi, D.~Xu, P.~H. Torr, and N.~Sebe, ``Edge guided gans with
  semantic preserving for semantic image synthesis,'' \emph{arXiv preprint
  arXiv:2003.13898}, 2020.

\bibitem{zhao2017pyramid}
H.~Zhao, J.~Shi, X.~Qi, X.~Wang, and J.~Jia, ``Pyramid scene parsing network,''
  in \emph{CVPR}, 2017.

\bibitem{he2019adaptive}
J.~He, Z.~Deng, L.~Zhou, Y.~Wang, and Y.~Qiao, ``Adaptive pyramid context
  network for semantic segmentation,'' in \emph{CVPR}, 2019.

\bibitem{hou2020strip}
Q.~Hou, L.~Zhang, M.-M. Cheng, and J.~Feng, ``Strip pooling: Rethinking spatial
  pooling for scene parsing,'' in \emph{CVPR}, 2020.

\bibitem{huang2019ccnet}
Z.~Huang, X.~Wang, L.~Huang, C.~Huang, Y.~Wei, and W.~Liu, ``Ccnet: Criss-cross
  attention for semantic segmentation,'' in \emph{ICCV}, 2019.

\bibitem{he2016deep}
K.~He, X.~Zhang, S.~Ren, and J.~Sun, ``Deep residual learning for image
  recognition,'' in \emph{CVPR}, 2016.

\bibitem{kingma2014adam}
D.~P. Kingma and J.~Ba, ``Adam: A method for stochastic optimization,'' in
  \emph{ICLR}, 2015.

\bibitem{paszke2019pytorch}
A.~Paszke, S.~Gross, F.~Massa, A.~Lerer, J.~Bradbury, G.~Chanan, T.~Killeen,
  Z.~Lin, N.~Gimelshein, L.~Antiga \emph{et~al.}, ``Pytorch: An imperative
  style, high-performance deep learning library,'' in \emph{NeurIPS}, 2019.

\bibitem{heusel2017gans}
M.~Heusel, H.~Ramsauer, T.~Unterthiner, B.~Nessler, and S.~Hochreiter, ``Gans
  trained by a two time-scale update rule converge to a local nash
  equilibrium,'' in \emph{NeurIPS}, 2017.

\bibitem{zhang2018unreasonable}
R.~Zhang, P.~Isola, A.~A. Efros, E.~Shechtman, and O.~Wang, ``The unreasonable
  effectiveness of deep features as a perceptual metric,'' in \emph{CVPR},
  2018.

\bibitem{qi2018semi}
X.~Qi, Q.~Chen, J.~Jia, and V.~Koltun, ``Semi-parametric image synthesis,'' in
  \emph{CVPR}, 2018.

\bibitem{li2020gan}
M.~Li, J.~Lin, Y.~Ding, Z.~Liu, J.-Y. Zhu, and S.~Han, ``Gan compression:
  Efficient architectures for interactive conditional gans,'' in \emph{CVPR},
  2020.

\bibitem{li2020bachgan}
Y.~Li, Y.~Cheng, Z.~Gan, L.~Yu, L.~Wang, and J.~Liu, ``Bachgan: High-resolution
  image synthesis from salient object layout,'' in \emph{CVPR}, 2020.

\bibitem{yu2017dilated}
F.~Yu, V.~Koltun, and T.~Funkhouser, ``Dilated residual networks,'' in
  \emph{CVPR}, 2017.

\end{thebibliography}
